\documentclass[lettersize,journal]{IEEEtran}
\usepackage[T1]{fontenc}
\usepackage{graphicx}
\usepackage{subcaption}
\usepackage{color}
\usepackage{textcomp}
\usepackage{enumitem}
\usepackage{csquotes}
\usepackage{hyperref}
\usepackage{amsmath} % for aligned environment

\usepackage{caption} % 调节标题的距离
\usepackage{tabularx}
\usepackage{float}
\usepackage{array}  % 对齐
\usepackage{booktabs} %调整表格线与上下内容的间隔
\usepackage{multirow}
\usepackage{makecell}
\usepackage{arydshln} % 负责画虚线的包
\usepackage{todonotes}
\usepackage{eurosym}
\usepackage{subcaption}
\usepackage{balance} % F: added this to balance the end

\usepackage{siunitx}
\usepackage{cite}
\begin{document}

% \title{Convey Emotions and Touch Gestures to Robots: Multimodal Decoding on Emotions and Social Touch Gestures.}

% \title{From Constrained to Free: Mapping Embodied Affective Touch Strategies on a Humanoid Robot}

\title{Mapping Embodied Affective Touch Strategies on a Humanoid Robot}

\author{Qiaoqiao Ren$^{13}$\and,
Omar Eldardeer$^{2}$\and,
Francesca Cocchella$^{2}$\and,
Rea Francesco$^{2}$\and, \\
Alessandra Sciutti$^{2}$\and,
and Tony Belpaeme$^{1}$

        % <-this % stops a space
\thanks{Faculty of Engineering and Architecture, $^{1}$IDLab-AIRO, Ghent University -- imec, Technologiepark 126, 9052 Gent, Belgium, $^{2}$ COgNiTive Architecture for Collaborative Technologies Unit of the Italian Institute of Technology, $^{3}$School of Electrical Engineering and Computer Science, KTH Royal Institute of Technology.}% <-this % stops a space

\thanks{The authors acknowledge the use of generative AI in preparing this manuscript. Specifically, Grammarly and GPT-5.2 were used to assist with grammar checking and enhancing the overall readability. All content was subsequently reviewed and edited by the authors, who accept full responsibility for the final version of the manuscript.}

}

\maketitle

\begin{abstract}

Affective touch in human-robot interaction is shaped not only by emotional intent, but also by the robot's embodiment, including where on the robot people touch, the physical constraints imposed by the robot body, and the robot's perceived agency and social role. Existing HRI studies, however, typically focus on only one or two isolated body parts, making it difficult to understand how affective touch generalizes across a full humanoid body. We present a study of embodied affective touch expression with 32 participants interacting with the iCub robot, a humanoid platform equipped with distributed tactile sensors across its full body. Participants expressed eight emotions under three conditions: unrestricted free touch, arm-only touch, and torso-only touch. The results show that 1). body region and spatial constraints jointly shape both where people touch and how they touch. In free touch, participants preferentially used socially accessible upper-body regions such as the torso and arms, whereas less frequently touched regions, such as the face/head and back, carried stronger emotion-specific selectivity. 2). Body region also shaped touch dynamics: in the arm-only condition, emotion-related variation was more evident in motion features, whereas in the torso-only condition it was more evident in pressure features. 3). Moreover, even for the same coarse body region, touching strategies did not transfer directly from free touch to the matched constrained condition, indicating that spatial constraint changes not only location choice but also the dynamics of touch expression. 4). Participants also reported greater closeness to the robot after expressing affective touch, and around 30\% reported a change in their perceived social relationship with the robot even in this one-directional, task-oriented interaction, highlighting the social impact of human-robot tactile interaction. Interaction analyses further showed that pressure patterns, more than motion patterns, were reconfigured by body region. Together, these findings demonstrate that affective touch strategies are body-region dependent and cannot be directly transferred across robot locations without considering the spatial constraints.

\end{abstract}

\section{Introduction}

Touch is one of the foundational and emotionally expressive modalities in human communication \cite{gallace2010science}. In human–human interaction, affective touch communicates care, reassurance, playfulness, empathy, dominance, and intimacy, often more efficiently, although  more ambiguously, than speech or facial expressions \cite{olugbade2023touch}. Touch is not merely a physical event; it is a biobehavioural, affective, and relational signal deeply shaped by context, history, and the bodies of the interactants \cite{suvilehto2023and}. Neuroscientific research shows that affective touch engages the C-tactile system and recruits neural circuits responsible for reward, social bonding, and emotion regulation \cite{kidd2023affective, olugbade2023touch}. Meanwhile, social psychology and anthropology demonstrate that touch behaviour reflects cultural norms, social roles, perceived vulnerability, and interpersonal dynamics \cite{suvilehto2023and, gallace2010science}. These insights illustrate a fundamental truth: touch is embodied, meaning that its meaning emerges from the interaction between two bodies, including how those bodies are perceived \cite{serino2010touch, gallese2013embodied}.

In human–robot interaction (HRI), touch holds enormous potential \cite{argall2010survey}. Robots are increasingly deployed in domains where emotional engagement and physical support matter—early education \cite{belpaeme2018social}, mental health care \cite{rabbitt2015integrating}, elder care \cite{abdollahi2022artificial}, child interaction \cite{vollmer2018children}, rehabilitation, and collaborative daily assistance \cite{mohebbi2020human}. Haptic technologies have advanced rapidly, enabling robots to sense contact with artificial skin, soft materials \cite{zhu2022soft, yin2021wearable}, distributed tactile arrays \cite{ren2024conveying, ren2025touch, hu2023can, dakopoulos20072d}, and force-sensitive surfaces \cite{tai2022force}. Yet despite this technological progress, touch remains one of the most underdeveloped and poorly understood modalities in HRI \cite{olugbade2023touch}. The majority of affective-touch research focuses on predetermined gestures performed on isolated body regions, such as stroking the forearm or tapping the hand \cite{teyssier2020conveying, zheng2019kinds}. These experimental constraints arise from practical limitations, robots possessing only small tactile patches or researchers defining gestures according to fixed sensor placement. The result is a simplified, gesture-centric view of tactile emotion expression that overlooks how humans naturally touch another embodied agent.

A growing body of work suggests that people do not touch robots the same way they touch humans \cite{olugbade2023touch}, objects \cite{ren2024tactile}, or animals \cite{sefidgar2015design}. Perceptions of the robot’s morphology, softness, childlikeness, fragility, or mechanical nature shape both the willingness to touch and the type, intensity, and location of touch \cite{argall2010survey}. Touching a small plush robot elicits different strategies compared to touching a tall humanoid robot with jointed limbs and a defined torso \cite{li2017touching, pu2020people}. Furthermore, studies demonstrate that touch is modulated by the robot’s perceived agency and intentional stance, whether the robot is thought to feel, react, or possess emotional states \cite{wiese2017robots, kwak2013makes}. A robot perceived as socially aware may invite gentle, affective contact, while a robot perceived as purely mechanical may evoke instrumental touch, avoidance, or even playful aggression \cite{ren2024tactile, van2013touch}. The robot’s social role, such as caregiver, care-receiver, assistant, patient, peer, or companion—further alters the boundaries of acceptable touch \cite{groom2011responses, lim2021social}, the emotions expressed, and the comfort level of the participant. These findings underscore a critical principle: robots are not merely sensory systems; they are experienced as social bodies \cite{parviainen2019motions}, whose form, materiality, agency, and role shape the emotional meaning and physical characteristics of human touch \cite{alavc2016social}.

Despite this evidence, the field lacks systematic data on how robot embodiment shapes affective touch expression. Existing datasets like \textit{Corpus of Social Touch} and \textit{The Human-Animal Affective Robot Touch} focus on one body part (usually on the forearm) and on specific touch gestures \cite{jung2015touch}, which may not generalize to other contexts according to embodiment theory \cite{borghi2015embodiment}. Touch strategies are shaped by the agent’s form, meaning that gestures expressed on a forearm may not translate to a torso, and strategies used on one robot cannot be assumed to generalize to another robot with a different morphology, material, or appearance. Without understanding these dependencies, models trained on limited gesture datasets risk misinterpreting emotional intent, while tactile sensor designs may miss regions that humans intuitively use for emotional communication. This absence of embodied knowledge also limits the design of robots capable of responding to touch in socially appropriate, relationally meaningful ways.

To address these gaps, we present a study of how people express affective touch toward a robot under both free full-body interaction and constrained body-part conditions. In contrast to prior research limited to specific sensor patches or experimenter-defined gestures, our study examines how participants express eight emotions across the robot’s body, and how these strategies shift when touch is spatially restricted (e.g., arm-only or torso-only). This allows us to capture both relatively unconstrained affective-touch behaviour and strategically adapted behaviour under spatial constraints, revealing how humans reorganize touch when only part of the robot body remains available. Beyond behavioural data, we incorporate measures of participants’ perceptions of the robot’s agency, emotional capacity, social role, and interpersonal closeness.

% Our findings highlight that affective touch is fundamentally embodied and context-dependent: emotional touch patterns vary not only by emotion but also by robot body region and by the participant’s interpretation of the robot’s role. Certain emotions cluster naturally around specific body areas, while others are reorganized when the available body region is constrained. These results demonstrate decisively that affective touch strategies are not universal and not transferable across robots or body regions without considering embodiment factors. This implies that effective affective-touch systems must model the robot’s full-body form, social identity, and perceived agency alongside tactile sensing, and cannot rely solely on abstract gesture taxonomies.

\section{Related work}

\subsection{Affective touch and social meaning in human–human interaction}

A rich body of research on human–human affective touch shows that touch is a powerful channel for communicating discrete emotions and affective qualities such as comfort, sympathy, love, anger, and gratitude \cite{hertenstein2006touch}. More recent analyses of physical interactions have revealed that affective touch involves distributed, time-varying pressure patterns and motion trajectories that often unfold across extended body areas rather than isolated points \cite{schirmer2021more, andreasson2018affective}. These findings support a view of touch as an embodied, relational behaviour shaped by both the toucher’s intention and the receiver’s body---its size, posture, perceived vulnerability, and social status \cite{remland1995interpersonal}. Previous research in social psychology and anthropology further shows that norms of ``body accessibility'' govern which body regions can be touched, by whom, and in what context, and that identical tactile sensations may be perceived as caring or invasive depending on relationship and situation \cite{sorokowska2021affective}. This provides a conceptual foundation for studying affective touch in HRI; if human touch patterns and their meanings are tightly coupled to the other’s body and social role, then robot touch cannot be reduced to body-part-agnostic sensor signals.

\subsection{Affective touch in human–robot interaction}

Within HRI, several seminal studies have adapted human–human tactile communication paradigms to robots. Andreasson, Alenljung, and colleagues used the NAO robot to investigate how humans convey eight discrete emotions via touch, largely on the robot’s arms and torso, and showed that people can systematically use tactile gestures to communicate both positive and negative emotions to a small humanoid \cite{andreasson2018affective}. Olugbade \textit{et al.} surveyed affective touch technology across human–human, human–robot, and virtual-human settings, concluding that affective touch can support comfort, bonding, and emotional regulation, but that existing systems tend to be restricted in spatial coverage and gesture diversity \cite{olugbade2023touch}. Some researchers have focused on the effects of being touched by a robot; for example, Willemse and colleagues showed that robot-initiated touch can attenuate physiological stress responses and increase perceived intimacy, suggesting that touch can be a valuable addition to social robots’ non-verbal repertoire \cite{willemse2017affective}. Reviews of touch technology in HRI and HCI consistently highlight the promise of tactile interaction but also stress that most studies use constrained, localised interfaces, making it difficult to generalise to full-body or ecologically complex scenarios \cite{ahmadpour2025affective, olugbade2023touch}. More recently, researchers have explored active social touch, where users initiate and control touch towards a robot. Our previous work showed that tactile interaction with a robot can influence people's behaviour and attitudes \cite{ren2024tactile}. Similarly, Gamboa-Montero \textit{et al.} showed that actively touching a humanoid robot can influence attitudes and engagement, and that the robot’s expressive behaviour during touch modulates these effects \cite{gamboa2025evaluating}. However, even in these studies, touch is typically limited to a small set of body parts and does not systematically examine how people would distribute emotional touch across a fully tactile robot body.

\subsection{Tactile gesture and emotion recognition in HRI}

On the decoding side, a substantial body of work has tackled tactile gesture recognition for social robots. Early efforts such as the \textit{Corpus of Social Touch} and \textit{the Social Touch Gesture} used wearable or matrix sensors to capture a variety of social touch gestures, and lots of researchers trained classifiers to recognise gesture categories from spatiotemporal pressure patterns \cite{jung2015touch}. In addition, Burns \textit{et al.} extended this approach by recognising not only gesture type but also force intensity, demonstrating that intensity carries additional social meaning and should be jointly modelled. Zhang \textit{et al.} did a survey about all the emotion recognition using affective touch \cite{zhang2025emotion}, and Gamboa-Montero \textit{et al.} proposed methods to detect, localise, and recognise human touches on a robot body, using distributed tactile sensors to infer both contact location and gesture type. Parallel work has begun to explore emotion recognition from touch. Textile-based pressure mapping systems and flexible electronic skins have been used to classify emotional states based on tactile patterns on a robot surface, sometimes integrating deep learning models to decode user emotion directly from tactile data \cite{pang2022skin}. Recent studies also extend decoding beyond tactile signals alone: audio-only models have been trained to recognise touch gestures and emotional arousal/valence from the sounds produced during contact with a robot, addressing platforms that lack full-body tactile sensing \cite{shi2022touching}\cite{shi2022touching}. 

Multimodal approaches combine touch, sound, and contextual information to improve robustness of emotion decoding in HRI. Despite these advances, most recognition systems are trained on data collected from one or two body regions, often with strictly defined gestures, and typically ignore how participants’ perception of the robot’s body, agency, and social role shapes the tactile expression itself. This raises the critical question of transferability: do touch strategies generalize across different robot body parts?

\subsection{``Body accessibility'' of human initiated touch in HRI}

In human-human tactile interaction, Suvilehto \textit{et al.} show that the extent of body areas people allow others to touch scales linearly with emotional closeness. These spatial patterns of social touch are consistent across cultures and reflect a fundamental mechanism for encoding and maintaining human social relationships \cite{suvilehto2015topography}. A smaller but growing literature explicitly examines body regions and full-body aspects of human-robot touch. Li et al. brought the concept of body accessibility into HRI by showing that tactile contact with different body parts of a humanoid robot elicits different levels of physiological arousal, echoing human--human norms about which body regions are more socially accessible \cite{li2017touching}. Follow-up work further showed that both body-part accessibility and anthropomorphic framing of robot body parts modulate physiological arousal during human-initiated touch \cite{maj2023touching}. Recent active-touch studies have also begun to examine how people distribute touch across a robot body in more naturalistic settings, rather than restricting interaction to one or two predefined contact areas \cite{andreasson2018affective}\cite{ren2024tactile}. In affective tactile HRI specifically, Andreasson \textit{et al.} showed that participants used varied touch patterns to convey emotions to a humanoid robot, underscoring that the robot body can function as an affective canvas rather than merely as a single contact site \cite{andreasson2018affective}. Beckerle and colleagues further argued that realistic social-touch interfaces should move beyond sparse local point contacts and instead support richer spatiotemporal and larger-surface tactile interaction \cite{beckerle2018feel}.

At the level of robotic hardware, platforms such as iCub, which offer full-body tactile coverage, have been used to study self-touch, body-model calibration, and visuo-tactile safety representations around the body \cite{roncone2014automatic}\cite{roncone2016peripersonal}. However, even in these systems, there is still little work on systematic affective emotion expression across all body regions, and on how humans actually choose to touch such robots when asked to convey specific emotions.

\subsection{Embodiment, agency, and social roles}

Beyond where people touch robots, a crucial line of work examines how embodiment, anthropomorphic framing, and social roles modulate the meaning and experience of touch. Maj \textit{et al.} showed that anthropomorphic framing and robot gender influence physiological arousal in touch interactions with social robots, demonstrating that how a robot is presented changes the bodily impact of the same physical interaction \cite{maj2023touching}. Broader HRI research on social norms and roles shows that robot design, perceived agency, and assigned roles significantly shape what users consider appropriate behaviour, including touch. Lawrence \textit{et al.} provide a systematic review of social norms in HRI, emphasising that trust, acceptance, and comfort depend on aligning robot behaviour with human expectations about roles, status, and relational distance \cite{lawrence2025role}. In parallel, embodiment-oriented HRI and design research argues that robots should be understood as social bodies: their form, materiality, movement, and responsiveness all contribute to how human touch is experienced and interpreted \cite{okuda2022human}\cite{maj2023touching}.

In conclusion, this literature shows that affective touch in HRI is shaped by body region, embodiment, and social interpretation, but it still leaves three key gaps. First, it remains unclear how tactile sensor placement should be prioritised on a full humanoid body for affective communication. Second, we still know little about how touch dynamics change across body regions, and whether emotion-related touch strategies transfer across constrained and unconstrained body access. Third, the attitudinal consequences of one-directional, task-oriented affective touch toward a robot remain underexplored. To address these gaps, this paper investigates the following three research questions:

\begin{itemize}

% \item How is affective touch distributed across the body of a full-body humanoid robot, and which regions should therefore be prioritised for tactile sensing?

\item Where do people touch a robot’s body when expressing different emotions, and how can these patterns inform the design and placement of tactile sensing systems?

\item Will people vary their touch dynamics across different robot body locations, and do these dynamics remain consistent on the same body part under different spatial constraints?

\item How does task-oriented, one-directional affective touch toward a robot influence people’s attitudes and impressions of the robot?

\end{itemize}

\section{Methodology}

To explore how affective touch is expressed across different body regions of a humanoid robot under spatial constraints, we conducted a dataset collection campaign.

\subsection{Participants}

A total of 32 participants ($37.10 \pm 14.93$) were recruited from the local community via mailing lists. Twenty-nine participants were Italian, and the remaining three were living in Italy and able to understand Italian and experimental notices. Participants were screened to ensure no tactile impairments and provided written informed consent in accordance with institutional ethical guidelines at the Istituto Italiano di Tecnologia (IIT), and the research was approved by the Regional Ethical Committee. The experiment lasted about one hour per participant, and all participants were compensated with 10 euros. Two participants' datasets were excluded because of incomplete recordings due to technical malfunctions, leaving 30 participants for the final analyses reported.

\subsection{Experiment Design}

\begin{figure}[t]
\centering
    \centering
    \includegraphics[width=0.5\linewidth]{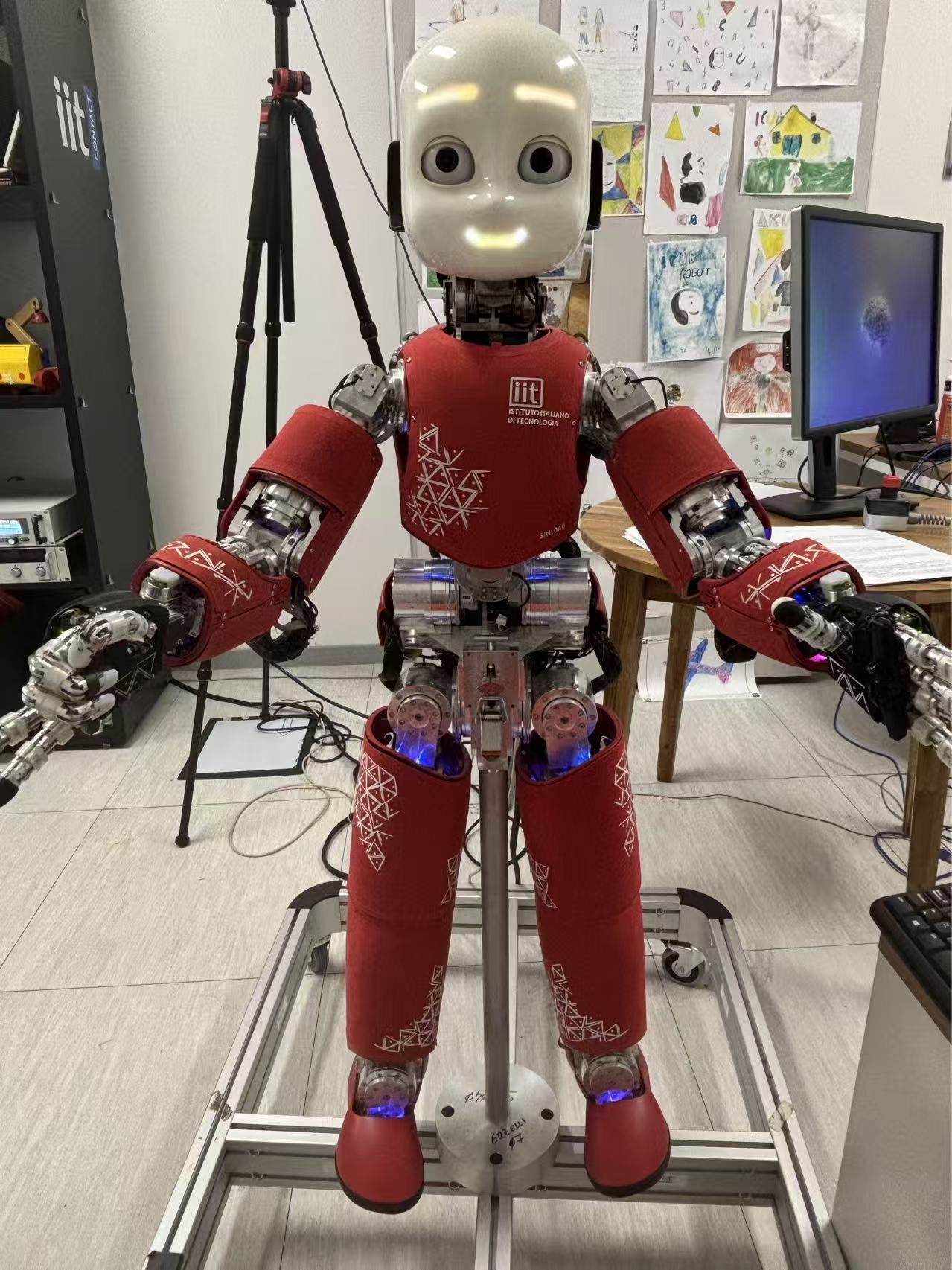}
    \caption{The iCub humanoid robot used in the study, equipped with distributed tactile sensing across the body. The red clothing covers the regions with tactile sensors. The robot is mounted on a movable base and further stabilised with a supporting rod to reduce the risk of falling.}
    \label{fig:icub_robot}
\end{figure}

\begin{figure*}[h]
\centering
\begin{subfigure}{0.28\textwidth}
    \centering
    \includegraphics[width=\linewidth]{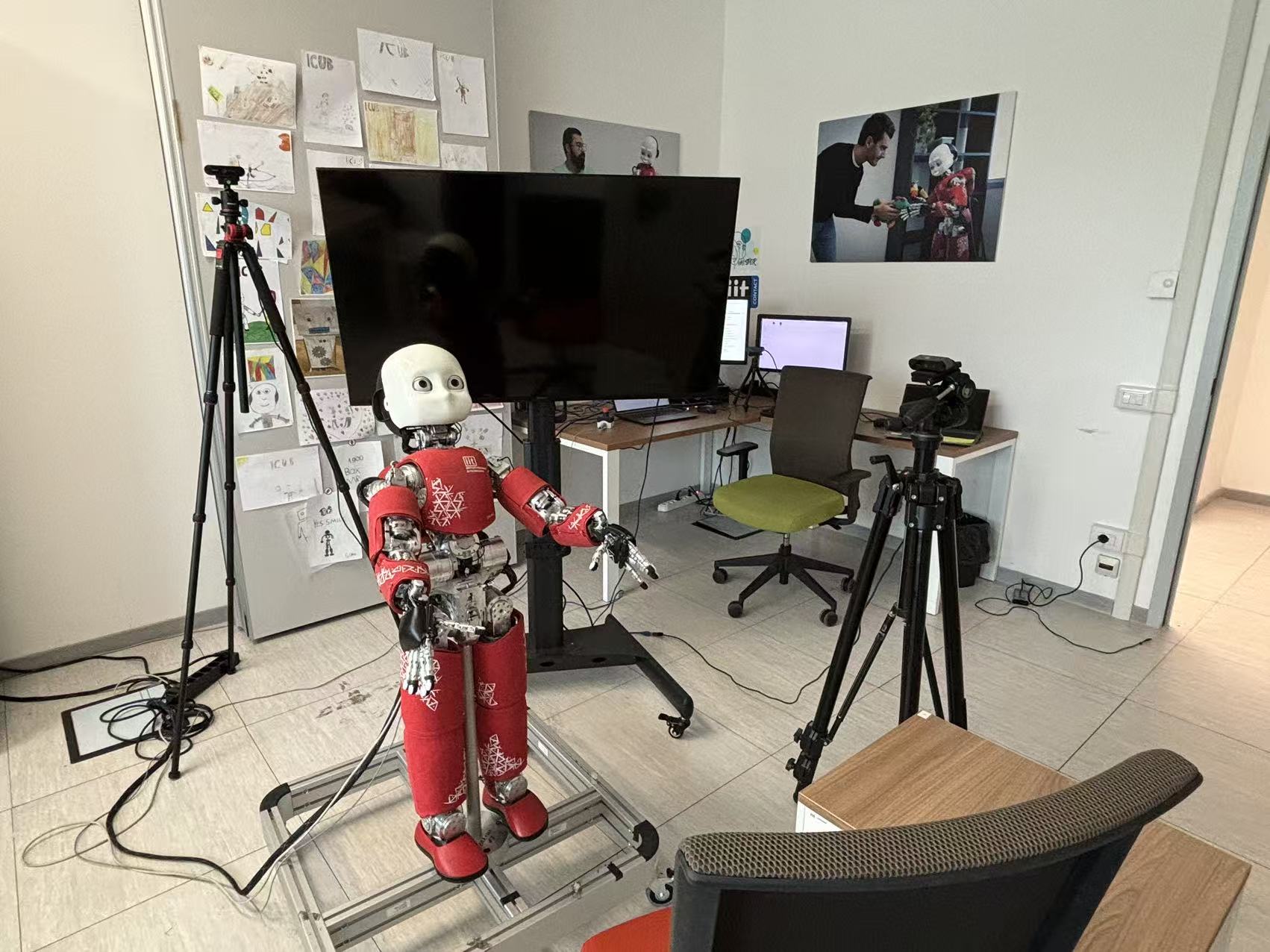}
    \caption{\parbox[t][1.3cm][t]{\linewidth}{Front view of the participant, showing the two cameras (camera 2 and camera 3) and the display screen.}}
    \label{fig:view1}
\end{subfigure}
\hspace{0.01\textwidth}
\begin{subfigure}{0.28\textwidth}
    \centering
    \includegraphics[width=\linewidth]{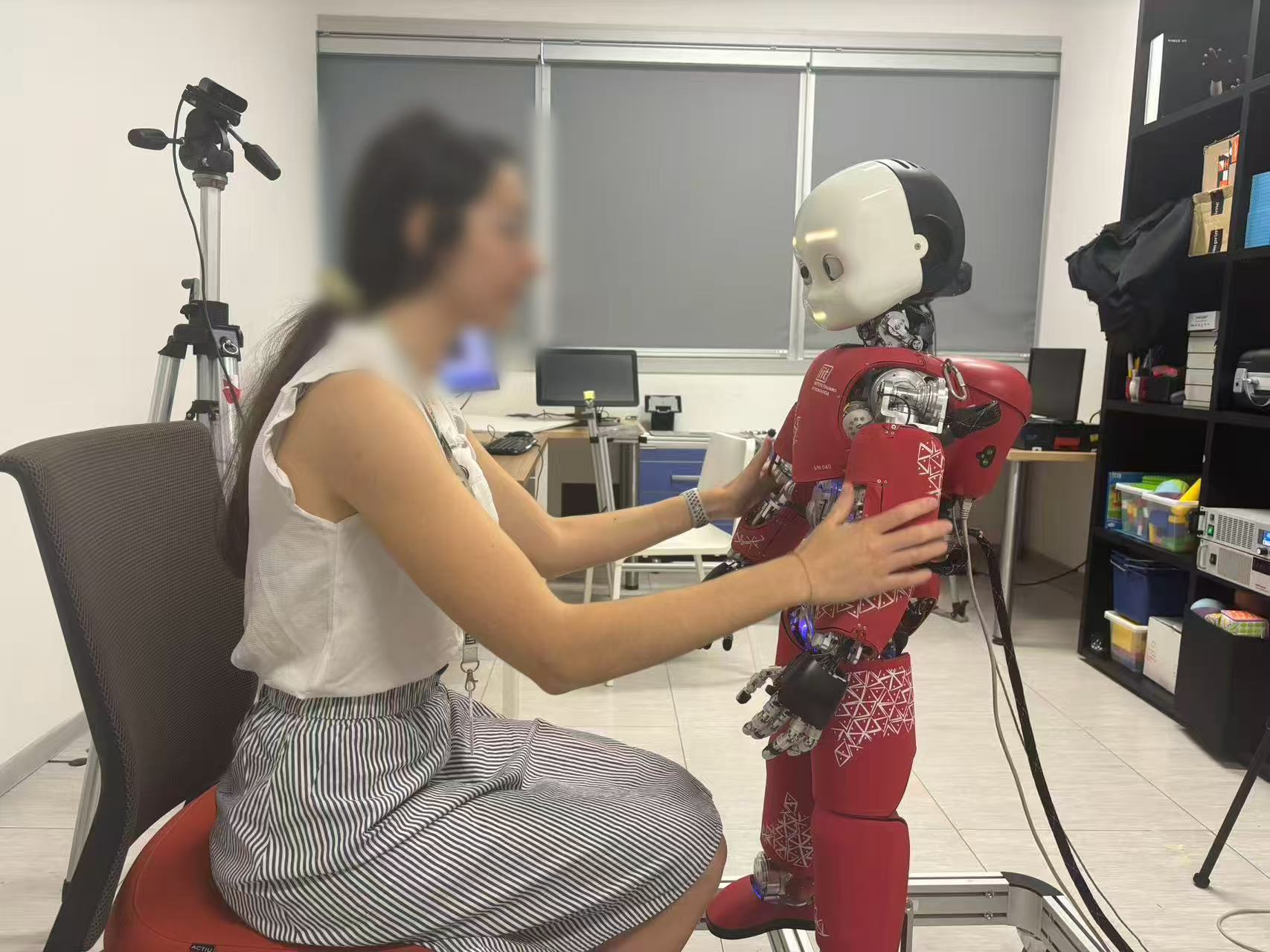}
    \caption{\parbox[t][1.3cm][t]{\linewidth}{Side view, showing one camera (camera 1) positioned behind, along with the participant’s posture and the robot’s posture.}}
    \label{fig:view2}
\end{subfigure}
\hspace{0.01\textwidth}
\begin{subfigure}{0.34\textwidth}
    \centering
    \includegraphics[width=\linewidth]{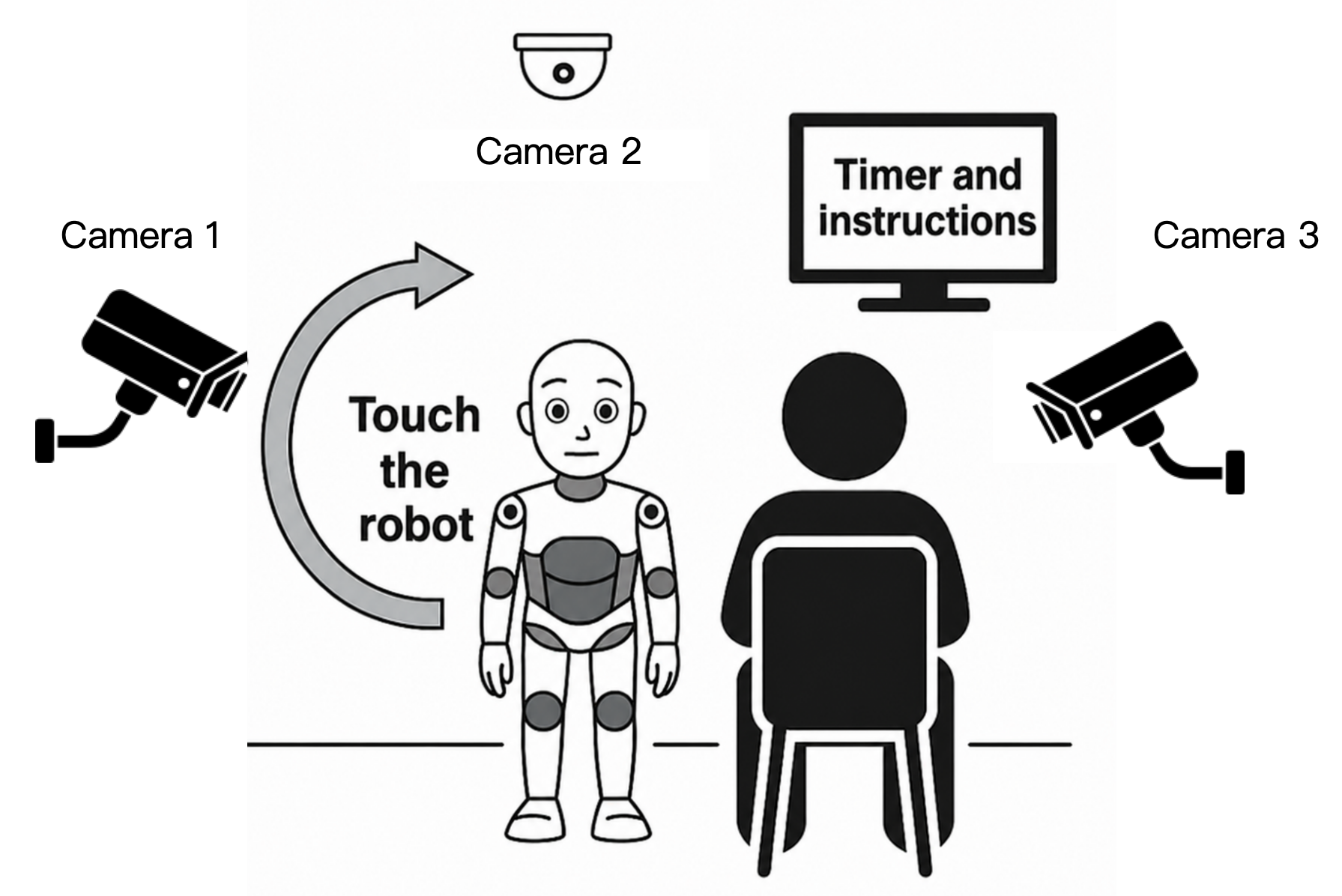}
    \caption{\parbox[t][1.3cm][t]{\linewidth}{Overview of experiment set-up}}
    \label{setup_chart}
\end{subfigure}
\caption{Experiment setup showing the front and side views of the participant, camera placements, and the overall configuration including the robot and display system.}
\label{fig:setupoverall}
\end{figure*}

Upon arrival, participants signed a consent form and completed the pre-questionnaire. They were then introduced to the iCub robot, as shown in Fig.~\ref{fig:icub_robot}, in an ``Introduction Phase'', where the robot makes eye contact, greets the participant with a wave, and explains the experiment rules using natural gestures. This priming was designed to reinforce the perception of the robot as a socially intentional agent. The experiment was conducted in a controlled setup (Fig.~\ref{fig:setupoverall}), where participants were seated facing the robot to reduce fatigue while the robot remained in a standing posture. The interaction was recorded using three external cameras (33 fps): one front camera, one side camera, and one rear camera capturing overall posture and movement. A display screen was positioned in front of the participant to present instructions and stimuli. Participants were free to stand up if needed to express certain touch interactions during the experiment.

Prior to the main experiment, a pre-session was conducted in which participants were allowed to freely explore the robot's body through touch from head to foot, with no restrictions on duration or body region, to reduce the novelty of the initial touch experience for participants. During the exploration, the robot was powered on but remained stationary with a neutral facial expression. The main experiment employed a within-subjects design featuring three touch conditions:

\begin{enumerate}
    \item Free Touch: Participants were free to choose any location on the robot’s whole body (from the head to the toe) to express the target emotion.
    \item Arm-Only Touch: Participants were instructed to use only the robot’s arm to convey the emotion.
    \item Torso-Only Touch: Participants were restricted to using only the robot’s torso during emotional expression.
\end{enumerate}

In the Free Touch condition, participants received no information about the location or density of tactile sensors, ensuring that touch behaviour was not guided by prior knowledge of the sensing system. For participants with prior exposure to the robot or who associated sensing with the robot’s clothing (skin), they were instructed to freely touch any part of the robot’s body, without regard to its sensing capabilities, and to focus on expressing the target emotion.

Each participant completed all three conditions, with the order of the \textit{Arm-Only} and \textit{Torso-Only} blocks counterbalanced across participants to mitigate order effects. The \textit{Free Touch} condition was always presented first to avoid priming participants with spatial restrictions.

% For all three conditions, participants were shown the name of a target emotion on the GUI. Following sensor calibration, a countdown was initiated, after which participants touched the robot for 10 seconds in accordance with the current condition. After each trial, the robot displayed a neutral facial expression and delivered a neutral verbal response to avoid emotional bias. Each emotion was expressed three times per condition, resulting in 24 trials per block. Trials were automatically advanced via the keyboard interface and GUI.

In each condition, participants expressed eight distinct affective states, following \cite{ren2024touch}, emotions were selected to cover different regions of the arousal–valence space, with one representative emotion per zone: Happiness (high arousal, positive valence), Anger (high arousal, negative valence), Sadness (low arousal, negative valence), and Comfort (low arousal, positive valence). In addition, inspired by \cite{hertenstein2006touch}, three socially oriented emotions were included: Love, Empathy, Gratitude; Emotions also include orienting responses and attention-related processes; we adopt a broad definition of emotion that includes interactional states such as ''grab attention” \cite{scherer2005emotions}, grab attention acts as a neutral affective state while some research also regard this as emotion \cite{hauser2019uncovering}. To avoid unnecessary terminological fragmentation, the present study uses the term 'emotions' as an umbrella label for these eight affective expressions but we do not treat these categories as strictly defined emotional states. Instead, they are used as representative affective and communicative interaction states, which include both emotional and social-intent signals. The goal of this study is not to compare emotions per se, but to examine how different interactional states are expressed through touch under embodiment constraints. Therefore, participants expressed eight emotions (\textit{Happiness, Sadness, Anger, Love, Empathy, Gratitude, Comfort, Grab attention}) across three repetitions, resulting in 24 trials per condition and 72 trials per participant.

In the main study, each trial proceeded as follows. The robot verbally instructed participants: “Please press [number] to express [emotion] when you are ready. (e.g., Please press 1 to express anger when you are ready.” After pressing the corresponding number, participants were shown the target emotion label on the GUI. A countdown was then displayed, during which participants touched the robot for 10 seconds according to the current condition. After each trial, the robot returned to a neutral facial expression and delivered a neutral verbal response (e.g., “Expression received, next one”) to minimize emotional carryover. Participants were informed that the robot would not respond to their actions during the experiment due to experimental constraints. Upon completion of all three blocks, participants completed a post-questionnaire and were debriefed. The full experimental configuration and source code are available on GitHub\footnote{\url{https://github.com/qiaoqiao2323/Affective_touch}}.

\begin{figure*}[h]
\centering
\begin{subfigure}{0.3\textwidth}
    \centering
    \includegraphics[width=\linewidth]{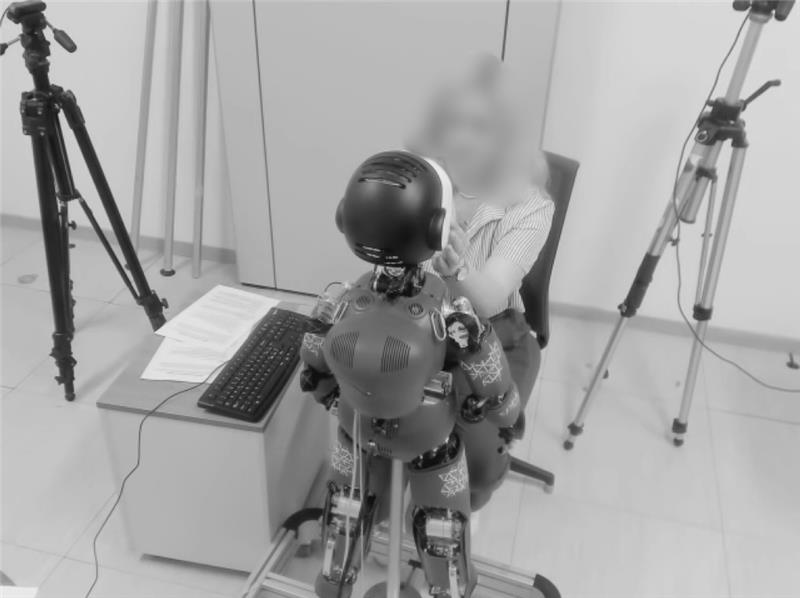}
    \caption{\parbox[t][1.2cm][t]{\linewidth}{Participant expresses ``love'' under the \textit{Free Touch} condition. Expression: holding the robot's face with both hands.}}
    \label{fig:free_2}
\end{subfigure}
\hspace{0.01\textwidth}
\begin{subfigure}{0.3\textwidth}
    \centering
    \includegraphics[width=\linewidth]{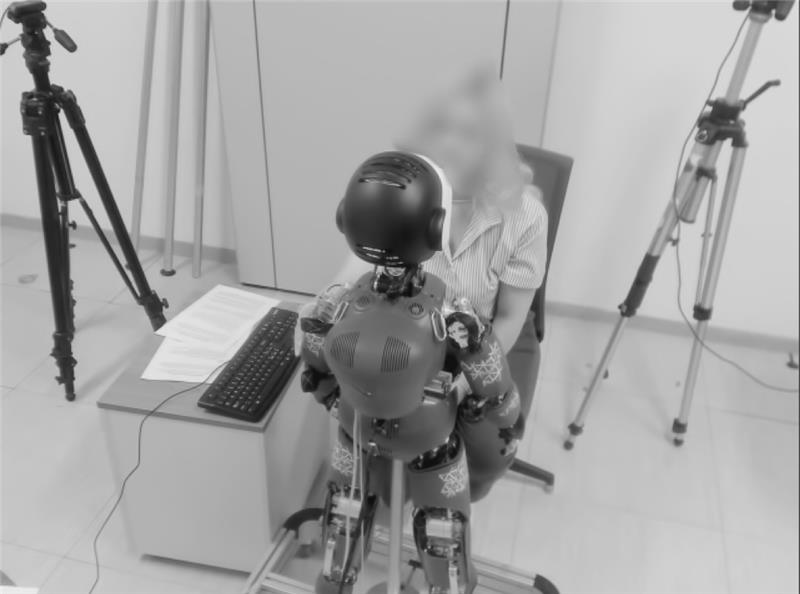}
    \caption{\parbox[t][1.2cm][t]{\linewidth}{Participant expresses ``love'' under the \textit{Arm\_Only Touch} condition. Expression: rubbing the robot's arm.}}
    \label{fig:arm_2}
\end{subfigure}
\hspace{0.01\textwidth}
\begin{subfigure}{0.3\textwidth}
    \centering
    \includegraphics[width=\linewidth]{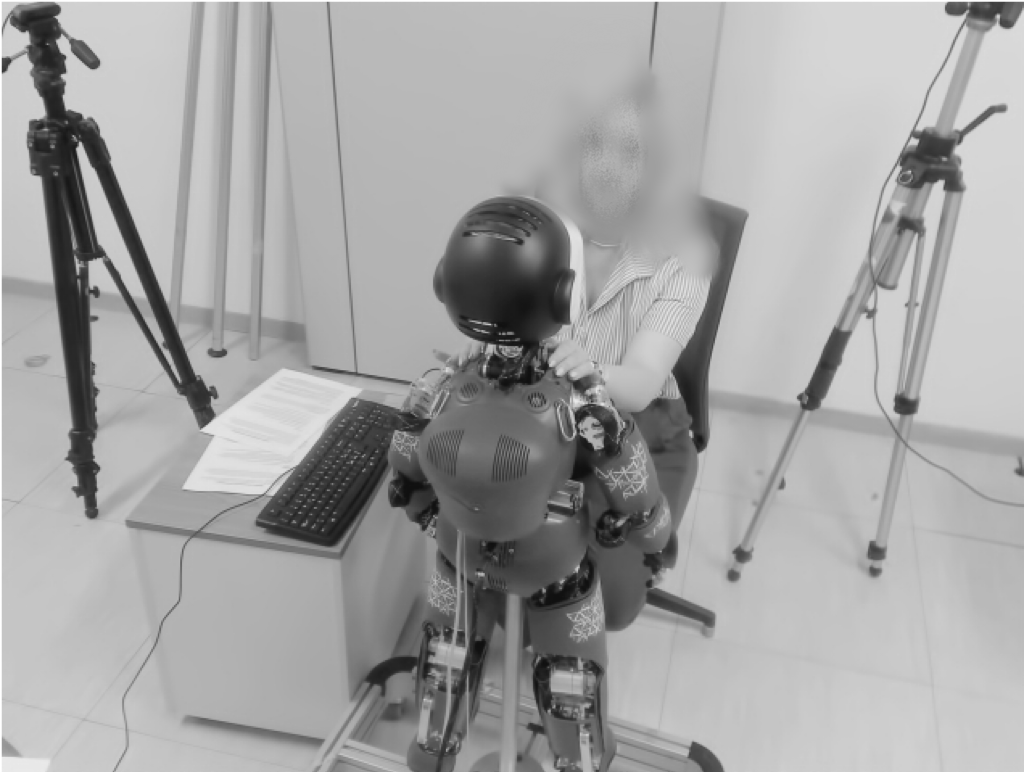}
    \caption{\parbox[t][1.2cm][t]{\linewidth}{Participant expresses ``love'' under the \textit{Torso\_Only Touch} condition. Expression: holding the upper torso near the shoulders with both hands.}}
    \label{fig:torso_2}
\end{subfigure}
\caption{The participant expressing ``love'' to the iCub robot.}
\label{fig:PB}
\end{figure*}

\subsection{Apparatus and Data Collection}

The experiment was conducted using the iCub humanoid robot developed by the Istituto Italiano di Tecnologia (IIT), a child-sized platform designed for embodied human–robot interaction. The robot is equipped with thirteen capacitive tactile patches distributed across the arms, torso, hands, legs, and feet. Each patch consists of multiple triangular sensing modules, forming a spatially distributed tactile sensing network. The number of taxels varies across body regions depending on surface area and sensor configuration. The taxel signals are continuous values representing contact intensity, with an 8-bit resolution. Detailed specifications of the tactile system are available in the official documentation\footnote{\url{https://robotology.github.io/docs/yarp/devices/skin/}}. 

% Participant interactions were recorded using three external high-resolution cameras (33 fps) capturing full-body behaviour, together with one built-in iCub facial camera used for face tracking. 

\subsubsection{Data Acquisition and Processing}
% Tactile Data Acquisition and Processing

Raw tactile data from the iCub skin were acquired via YARP datadumpers, where each record contains a timestamp and a vector of taxel readings. A keyboard-triggered control signal was used to synchronise all data streams and to mark trial onset and offset, enabling consistent segmentation across modalities. The raw logs were first parsed into frame-wise taxel vectors. Frames were retained only if at least one taxel exceeded a pressure threshold of 10, removing low-level sensor noise. Trial segments were extracted using the keyboard-triggered timestamps. For each trial, tactile signals within the corresponding time window were aggregated to form a single trial-level representation.

% Video Segmentation and Processing

All video streams were synchronised with tactile data using shared timestamps. Video sequences were segmented into trials based on the same keyboard-triggered markers used for tactile data. Within each trial, frames were resampled to 3 frames per second by selecting the temporally nearest frame to each target timestamp. All frames were converted to grayscale to reduce data dimensionality. Motion signals were computed from the segmented video using frame-difference energy. Prior to differencing, each frame was smoothed using a Gaussian filter. Pixel differences below a threshold of 3 were suppressed to reduce noise. The motion energy for each frame was defined as the mean absolute pixel difference between consecutive frames. Motion signals were standardised independently for each camera stream. Trial-level motion representations were obtained by aggregating frame-level motion values within each segmented trial window.

All tactile and video signals were aligned using shared timestamps and keyboard-triggered trial markers. Segmentation was applied consistently across modalities, ensuring that tactile and video data correspond to the same temporal intervals. All features were aggregated at the trial level for subsequent analysis.

\subsubsection{Pre-Session Questionnaire.}  
The pre-session questionnaire established baseline perceptions and expectations about the robot. Participants reported their demographics, including age, gender, nationality, and the highest completed education level. They answered the following scales.

\textbf{Previous experiences with robots.} They indicated their general familiarity with robots (e.g., ``I have never seen one before,''  ``I work with robots every day'')  and whether they had previously participated in an iCub study. A multiple-choice question asked where they had seen iCub before (e.g., on TV, on social media, in a laboratory, elsewhere, or never).
    
\textbf{Agency and Experience scale.} Eight statements evaluated the perceived capacity of iCub for acting and feeling. \cite{gray2007dimensions}. Items included: \textit{``I think iCub can feel pain,'' ``I think iCub can feel fear,'' ``I think iCub can feel pleasure,'' ``I think iCub can feel joy,'' ``I think iCub is capable of morality,'' ``I think iCub is capable of planning,'' ``I think iCub can recognize emotions,'' and ``I think iCub has self-control.''} Each was rated on a 7-point Likert scale (1 = Not at all agree, 7 = Totally agree).

\textbf{Agency and Communality traits.} Participants evaluated how well 11 adjectives described iCub.  \cite{fiske2007universal}
\textbf{Agency traits:} active, dynamic, efficient, assertive, confident.  
\textbf{Communality traits:} friendly, empathetic, trustworthy, understanding, helpful, likeable.  Ratings used a 7-point scale (1 = Not at all, 7 = Totally).

\textbf{Social role attribution.} Inspired by \cite{eyssel2012social}, participants selected which social role best described iCub for them, choosing from: ``student,'' ``relative,'' ``stranger,'' ``neighbour,'' ``friend,'' or ``colleague.''

\textbf{IOS: Inclusion of the Other in the Self, a measure of social and mental closeness.} Participants viewed seven pairs of overlapping circles and selected the one that best represented how close they felt to iCub \cite{aron1992inclusion}.

\subsubsection{Post-Session Questionnaire.}  
After interacting with iCub, participants completed a follow-up survey to capture shifts in perception and their emotional experience of the interaction. It included:

Participants re-rated the same eight \textit{Agency and Experience} statements and the same 11 \textit{Agency and Communality} traits from the pre-survey, allowing comparison of before-and-after perceptions. And participants were again asked which social role best described iCub after the interaction. In addition, participants were again asked to rate the IOS scale.

\textbf{Comfort ``thermometer'' \cite{alwin2007margins}} A visual-analog scale (0–100\%) asked participants how comfortable they felt when touching iCub, with 0\% indicating ``totally uncomfortable,'' 50\% ``neither comfortable nor uncomfortable,'' and 100\% ``totally comfortable.''

\section{Results}

\subsection{Spatial preferences and socially salient body regions}

For the spatial analysis, the robot body was partitioned into 21 predefined touch locations: Left Upper Arm, Right Upper Arm, Torso, Right Hand, Left Hand, Right Forearm, Left Exposed Joint (shoulder), Left Forearm, Face and Head, Right Exposed Joint (shoulder), Back, Left Leg Upper, Right Leg Upper, Waist, Right Leg Lower, Left Leg Lower, Neck, Left Foot, Right Foot, Pelvic Region, and Gluteal Region, as shown in Fig.~\ref{fig:heatmap}. We annotated the touched non-sensorized body regions using three camera views and combined these annotations with the tactile sensor data (sensorized body regions). The free-touch condition revealed a strongly non-uniform distribution of tactile choices across the robot body. The most frequently touched regions were the upper arms, torso, hands, forearms, exposed joints, and face/head, whereas the waist was only occasionally used, and the legs, feet, pelvic region, and gluteal region were almost never selected. Fig.~\ref{fig:overall_body_parts} summarizes touch frequency across body regions in the free-touch position matrix. These counts indicate that socially meaningful and physically accessible upper-body regions carried most of the affective touch expressions, while lower-body regions contributed little to spontaneous emotional expression. Figs.~\ref{fig:heatmap} visualise how frequently each body region was touched, highlighting preferred touch patterns. The dominance of the upper arms, torso, and hands is not a weak tendency but a steep frequency drop-off across the ranked body parts. To test whether body part choice depended on the expressed emotion, we fitted a separate mixed-effects logistic model for each predefined body part:

\begin{figure*}
\centering
\begin{subfigure}[t]{0.48\linewidth}
    \centering
    \includegraphics[width=\linewidth]{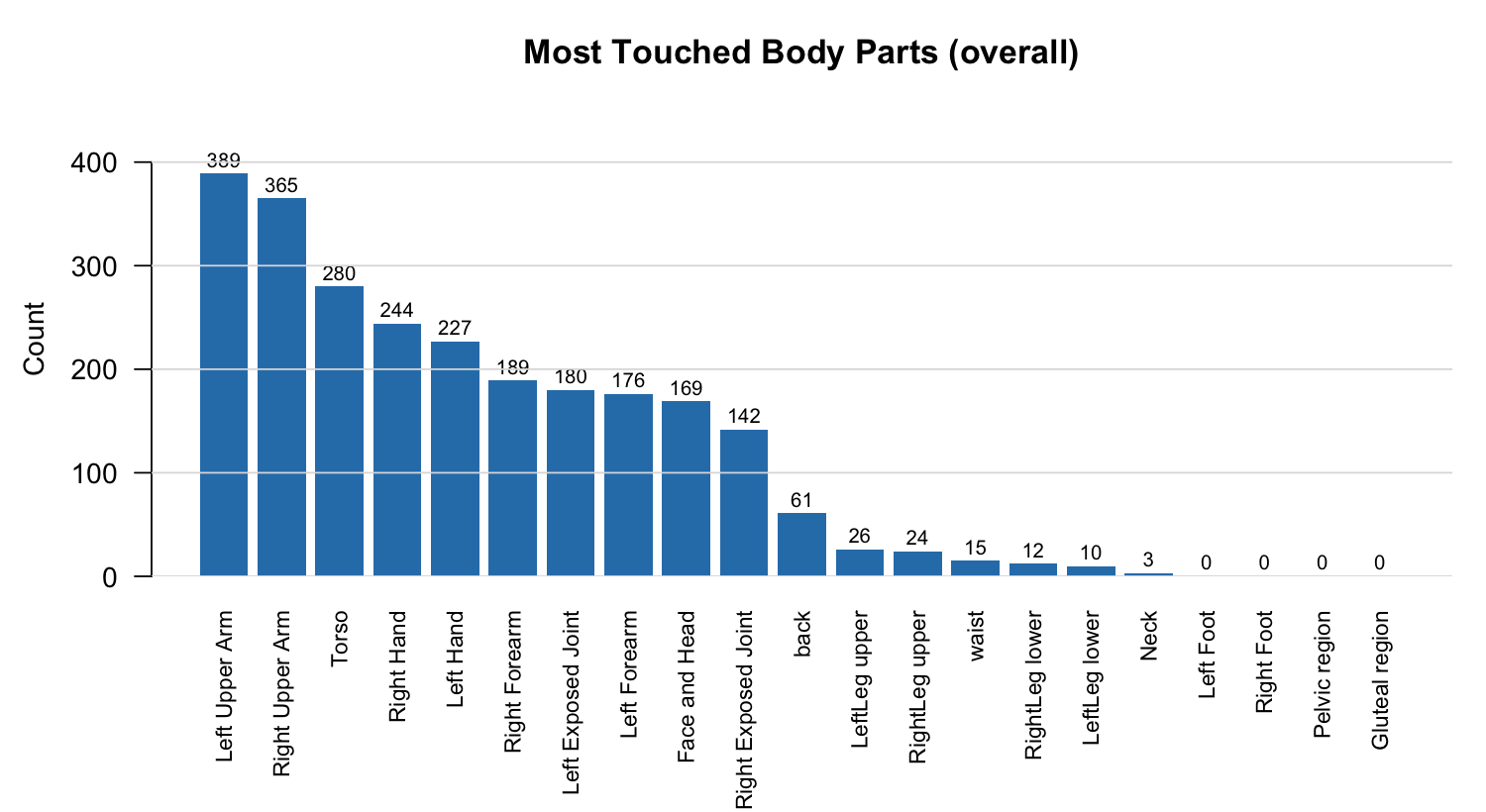}
    \caption{Overall frequency of touched body parts in the free-touch condition. Upper-body regions dominate spontaneous affective touch, while lower-body regions are rarely selected.}
    \label{fig:overall_body_parts}
\end{subfigure}
\hfill
\begin{subfigure}[t]{0.48\linewidth}
    \centering
    \includegraphics[width=\linewidth]{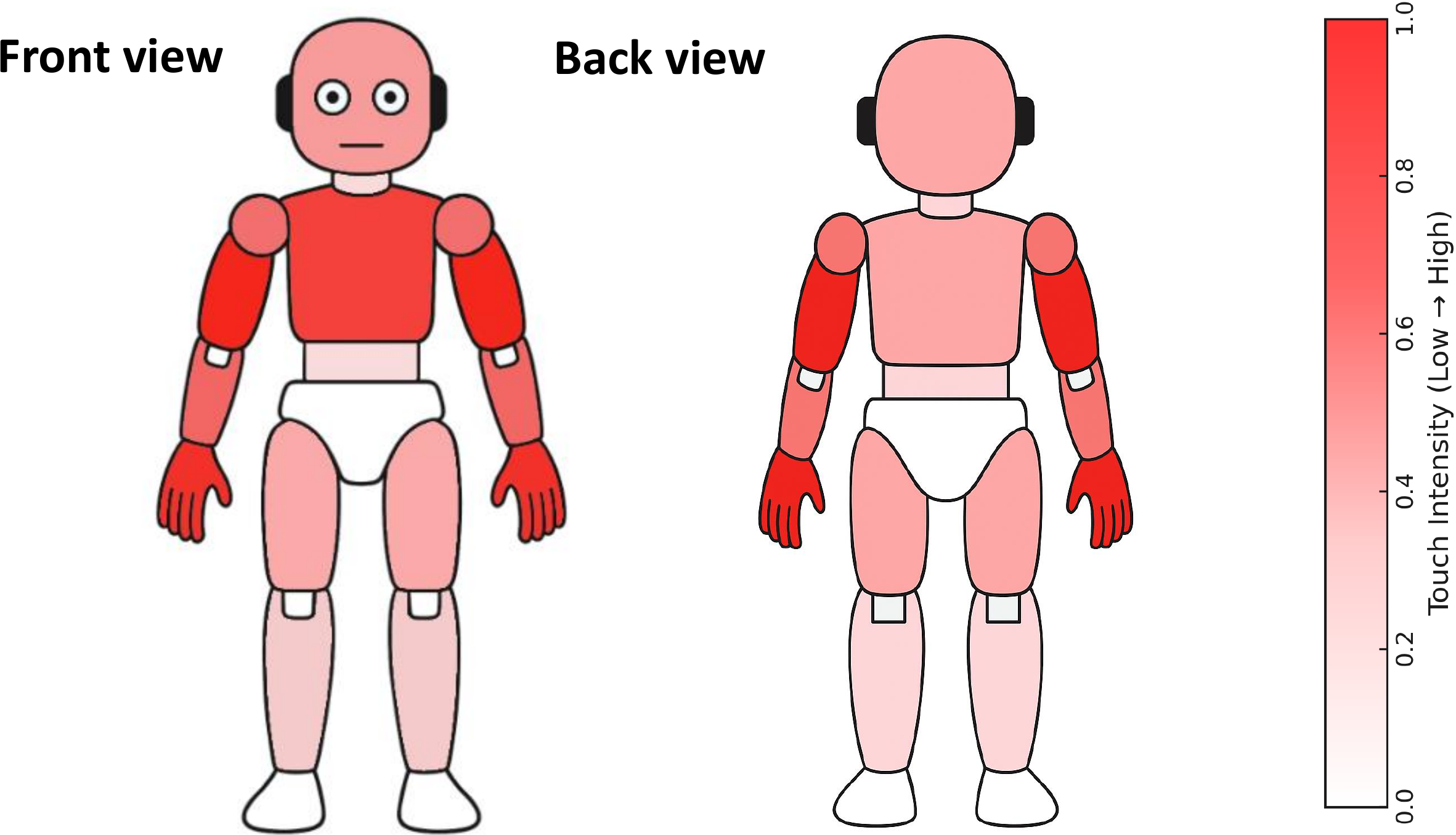}
    \caption{Heatmap of touch activity across body regions.}
    \label{fig:heatmap}
\end{subfigure}
\caption{Body acceptance of affective touch}
\label{fig:Body_acceptance}
\end{figure*}

\begin{equation}
\begin{aligned}
\text{logit}\!\left(P(\text{BP}_{ij}=1)\right)
&= \beta_0 + \beta_1\,\text{Emotion}_{ij} + u_j \\
u_j &\sim \mathcal{N}(0,\sigma_u^2)
\end{aligned}
\end{equation}

Here, $\text{BP}_{ij}$ indicates whether a given body part was touched on trial $i$ for participant $j$, \textit{Emotion} is the fixed effect, and $u_j$ is a participant-specific random intercept. For each body part, the full model was compared with a random-intercept-only null model using a likelihood-ratio test,

\begin{equation}
\chi^2 = 2\left(\ell_{\mathrm{full}}-\ell_{\mathrm{null}}\right).
\end{equation}

% Among the higher-frequency body parts, significant emotion effects were observed for the right hand ($\chi^2(7)=85.68$, $p<.001$), left hand ($\chi^2(7)=67.14$, $p<.001$), torso ($\chi^2(7)=53.38$, $p<.001$), left upper arm ($\chi^2(7)=42.86$, $p<.001$), face and head ($\chi^2(7)=42.86$, $p<.001$), right upper arm ($\chi^2(7)=24.22$, $p=.001$), and left exposed joint ($\chi^2(7)=16.54$, $p=.021$). Significant effects were also found for several lower-frequency body parts, including the back ($\chi^2(7)=62.90$, $p<.001$), left upper leg ($\chi^2(7)=40.78$, $p<.001$), right upper leg ($\chi^2(7)=34.93$, $p<.001$), right lower leg ($\chi^2(7)=25.38$, $p<.001$), left lower leg ($\chi^2(7)=24.86$, $p<.001$), waist ($\chi^2(7)=18.72$, $p=.009$), and neck ($\chi^2(7)=15.15$, $p=.034$). In conclusion, these results show that emotion systematically reorganized body-part choice across the robot body, rather than merely modulating touch dynamics within a fixed location.

Significant effects of emotion were observed across multiple body regions (Table~\ref{tab:chi_body_regions}), spanning both frequently and less frequently touched areas. Stronger effects were observed in commonly engaged regions such as the hands and torso, suggesting that these areas are more sensitive to emotional modulation. In contrast, less frequently touched regions (e.g., neck and waist) also showed significant effects, albeit with smaller magnitudes. Overall, these results indicate that emotion systematically influences the spatial distribution of touch across the body, rather than only modulating touch dynamics within fixed locations.

\begin{table}[t]
\centering
\caption{Chi-square test results for emotion effects across body regions.}
\label{tab:chi_body_regions}
\begin{tabular}{lcc}
\toprule
\textbf{Body Region} & $\boldsymbol{\chi^2(7)}$ & \textbf{p-value} \\
\midrule
Right hand          & 85.68 & $< .001$ \\
Left hand           & 67.14 & $< .001$ \\
Torso               & 53.38 & $< .001$ \\
Left upper arm      & 42.86 & $< .001$ \\
Face \& head        & 42.86 & $< .001$ \\
Right upper arm     & 24.22 & $= .001$ \\
Left exposed joint  & 16.54 & $= .021$ \\
\midrule
Back                & 62.90 & $< .001$ \\
Left upper leg      & 40.78 & $< .001$ \\
Right upper leg     & 34.93 & $< .001$ \\
Right lower leg     & 25.38 & $< .001$ \\
Left lower leg      & 24.86 & $< .001$ \\
Waist               & 18.72 & $= .009$ \\
Neck                & 15.15 & $= .034$ \\
\bottomrule
\end{tabular}
\end{table}

Among the higher-frequency body parts, significant emotion effects were observed for all regions examined, including the right hand ($\chi^2(7)=85.68$, $p<.001$), left hand ($\chi^2(7)=67.14$, $p<.001$), torso ($\chi^2(7)=53.38$, $p<.001$), left upper arm ($\chi^2(7)=42.86$, $p<.001$), face and head ($\chi^2(7)=42.86$, $p<.001$), right upper arm ($\chi^2(7)=24.22$, $p=.001$), and left exposed joint ($\chi^2(7)=16.54$, $p=.021$), with stronger effects generally observed in the hands and torso. Significant effects were also found across all lower-frequency body parts, including the back ($\chi^2(7)=62.90$, $p<.001$), left upper leg ($\chi^2(7)=40.78$, $p<.001$), right upper leg ($\chi^2(7)=34.93$, $p<.001$), right lower leg ($\chi^2(7)=25.38$, $p<.001$), left lower leg ($\chi^2(7)=24.86$, $p<.001$), waist ($\chi^2(7)=18.72$, $p=.009$), and neck ($\chi^2(7)=15.15$, $p=.034$), where effect sizes were comparatively smaller for regions such as the waist and neck. Overall, these results indicate that emotion systematically reorganized body-part choice across the robot body, rather than merely modulating touch dynamics within fixed locations.

These lower-frequency regions should be interpreted more cautiously at the level of emotion-specific preference because they were touched much less often overall. At the same time, their low frequency is itself meaningful rather than incidental, because it reflects the social accessibility of the robot body. In other words, the fact that some regions were rarely or never used is part of the embodied result: emotional touch was constrained not only by expressive preference, but also by what participants perceived as acceptable, reachable, or socially appropriate locations on the robot. For example, no participant selected the pelvic or gluteal region, even though the interaction target was a robot rather than a human. Frequent use therefore does not necessarily mean that a body part carries stronger emotional weight. To capture this distinction, we computed a centered logit effect for each emotion and body-part combination:

\begin{equation}
c_{eb} = \operatorname{logit}(\hat{p}_{eb}) 
- \frac{1}{E}\sum_{e'} \operatorname{logit}(\hat{p}_{e'b})
\end{equation}

where $\hat{p}_{eb}$ is the model-predicted probability of touching body part $b$ under emotion $e$, and $E$ is the number of emotions. This quantity expresses how strongly an emotion elevates or suppresses the use of a given body part relative to that body part's own mean usage across emotions. For each emotion, the body part with the largest $c_{eb}$ was defined as the most selectively associated body part.

\begin{figure*}[t]
\centering
\begin{subfigure}[t]{0.8\linewidth}
    \centering
    \includegraphics[width=\linewidth]{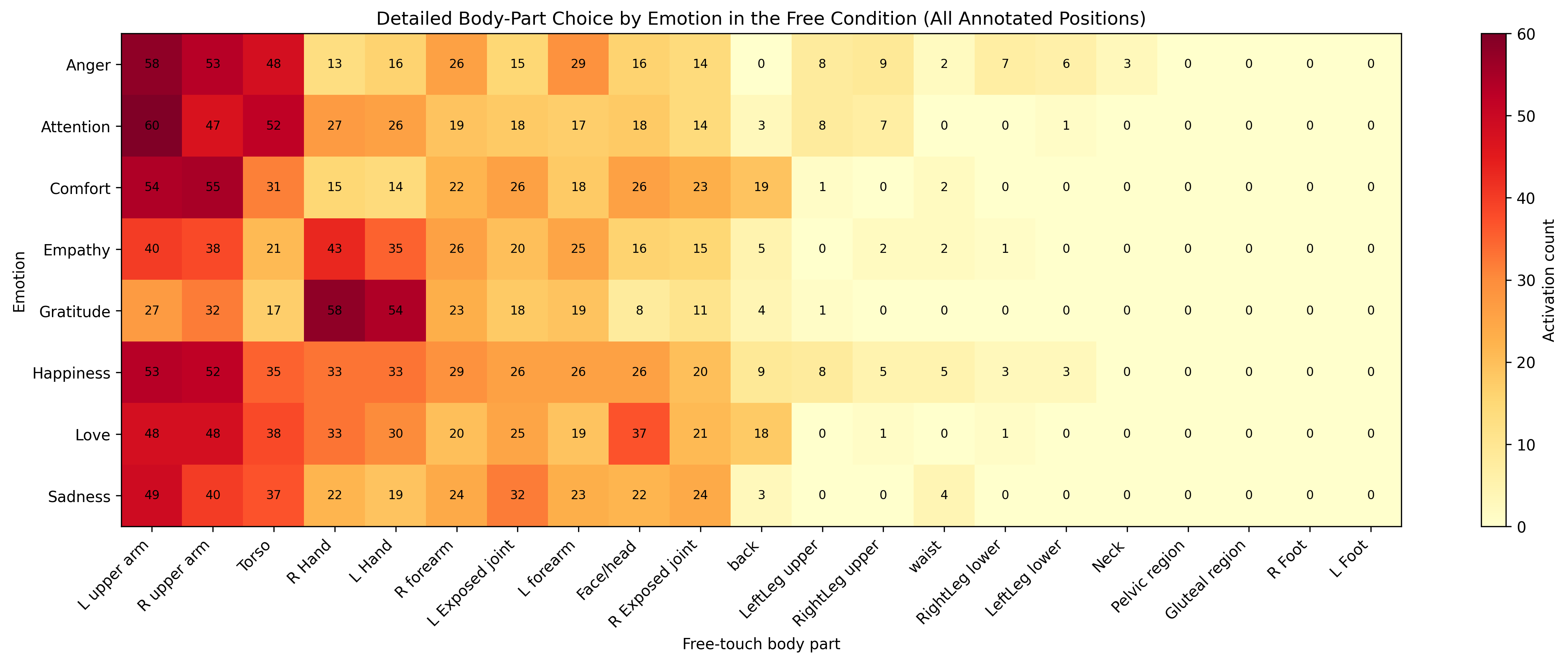}
    \caption{Detailed body-part choice by emotion in the free-touch condition.}
    \label{fig:free_body_location_emotion}
\end{subfigure}
% \hfill
% \begin{subfigure}[t]{0.8\linewidth}
%     \centering
%     \includegraphics[width=\linewidth]{figures/bodypart_emotion_probability_flow_normalized.png}
%     \caption{Body part normalized probability flow for free touch body part selection. Each body part has equal total width, so the flows represent the emotion composition within each body part rather than the absolute touch frequency. This representation allows emotion--body part associations to be compared more fairly across body parts with different overall usage rates.}
%     \label{fig:bodypart_emotion_flow_norm}
% \end{subfigure}
\caption{Body part distributions and normalized emotion body part associations in the free touch condition.}
\label{fig:combined_free_touch}
\end{figure*}

As shown in Figure~\ref{fig:free_body_location_emotion}, the plot reveals the internal emotion composition of each body part. Under this normalized comparison, the right hand is dominated by \textit{gratitude} and \textit{empathy}, the face and head are relatively enriched for \textit{love}, and the torso is enriched for \textit{anger} and \textit{attention}. The figure reflects selective preference rather than touch frequency. Table~\ref{tab:emotion_bodypart_summary} clarifies this distinction. In absolute terms, the upper arms dominate across most emotions, while \textit{gratitude} and \textit{empathy} shift toward the hands. In relative terms, lower-frequency body parts become informative because they are selectively overrepresented once baseline usage is removed. For example, \textit{comfort} and \textit{love} show their strongest selective association on the back, despite its low overall touch frequency. These results indicate that body acceptance does not directly map to emotional weight. This spatial pattern partially addresses the first research question; Sensor placement should therefore balance two factors. Frequently used body parts should be covered with higher sensitivity to capture common interaction patterns, while selectively informative regions should also be instrumented, even if they are less frequently touched. Notably, several such regions, including the face, exposed joints (shoulder), and the back, are not fully covered in the current iCub configuration.

\begin{table*}[t]
\centering
\caption{Model-derived summary of the strongest emotion-body-part links in the free-touch condition. ``Most likely body part (MLBP)'' refers to the highest average predicted touch probability;``Most selectively associated body part (MABP)'' refers to the largest centered logit effect. The table also reports the second-highest selectively associated body part (2nd MABP) for each emotion. Significance values refer to the overall body-part-specific mixed-effects logistic model for the listed body part, not to a pairwise test of the individual link.}
\label{tab:emotion_bodypart_summary}
\begin{tabular}{l l c c c l c c c l c c c}
\toprule
Emotion & MLBP & $\hat{p}$ & $\chi^2(7)$ & $p$ & MABP & $c_{eb}$ & $\chi^2(7)$ & $p$ & 2nd MABP & $c_{eb}$ & $\chi^2(7)$ & $p$ \\
\midrule
Anger & Left upper arm & 0.65 & 42.86 & $<0.01$ & Torso & 0.69 & 53.38 & $<0.01$ & Left upper arm & 0.51 & 42.86 & $<0.01$ \\
Attention & Left upper arm & 0.67 & 42.86 & $<0.01$ & Torso & 0.89 & 53.38 & $<0.01$ & Left upper arm & 0.62 & 42.86 & $<0.01$ \\
Comfort & Right upper arm & 0.61 & 24.22 & $<0.01$ & Back & 2.74 & 62.90 & $<0.01$ & Right upper arm & 0.49 & 24.22 & $<0.01$ \\
Empathy & Right hand & 0.46 & 85.68 & $<0.01$ & Back & 0.76 & 62.90 & $<0.01$ & Right hand & 0.73 & 85.68 & $<0.01$ \\
Gratitude & Right hand & 0.64 & 85.68 & $<0.01$ & Right hand & 1.48 & 85.68 & $<0.01$ & Left hand & 1.41 & 67.14 & $<0.01$ \\
Happiness & Left upper arm & 0.58 & 42.86 & $<0.01$ & Back & 1.53 & 62.90 & $<0.01$ & Face and head & 0.45 & 42.86 & $<0.01$ \\
Love & Right upper arm & 0.52 & 24.22 & $<0.01$ & Back & 2.64 & 62.90 & $<0.01$ & Face and head & 1.19 & 42.86 & $<0.01$ \\
Sadness & Left upper arm & 0.53 & 42.86 & $<0.01$ & Left shoulder & 0.67 & 16.54 & $0.021$ & Back & 0.16 & 62.90 & $<0.01$ \\
\bottomrule
\end{tabular}
\end{table*}

\subsection{Touch dynamics across free and constrained conditions}

After identifying which body regions were preferentially used and which carried greater emotional weight, we next examined how participants executed affective touch once a body region was available. Emotional intent could already be conveyed prior to contact through approach dynamics, such as movement speed, and further expressed through applied force. Importantly, faster movements did not necessarily correspond to higher pressure. The analysis therefore shifted from body-part selection to the temporal and physical dynamics of touch, including both pressure-related and motion-related patterns, and how these varied between free-touch and spatially constrained conditions.

Based on the collected tactile and video data, we selected a small set of representative pressure and motion features to support interpretable comparison across the three conditions rather than exhaustively using the full feature pool. The features were chosen to cover complementary aspects of touch, including contact extent, pressure intensity, pressure variation, and temporal dynamics, while reducing redundancy among highly correlated variables. The representative pressure features were \textit{Max Touch Area}, \textit{Mean Touch Duration}, \textit{Pressure Gradient}, \textit{Pressure Std}, \textit{Mean Pressure}, and \textit{Max Pressure}. The representative motion features were \textit{mean spectral centroid}, \textit{mean peak energy}, \textit{mean temporal centroid}, \textit{max peak energy}, \textit{mean total energy}, and \textit{mean energy}.

% \begin{figure*}[t]
% \centering
% \begin{subfigure}[t]{0.49\textwidth}
%     \centering
%     \includegraphics[width=\linewidth]{figures/figure_06_max_pressure_by_condition_emotion.png}
%     \caption{\textit{Max\_Pressure} comparison among torso-only, arm-only and free conditions.}
%     \label{fig:max_pressure_conditions}
% \end{subfigure}
% \hfill
% \begin{subfigure}[t]{0.49\textwidth}
%     \centering
%     \includegraphics[width=\linewidth]{figures/figure_07_motion_energy_by_condition_emotion.png}
%     \caption{Condition-level profiles for \textit{Mean\_Total\_Energy}.}
%     \label{fig:motion_conditions}
% \end{subfigure}
% \caption{Pressure and motion features across emotions and conditions. These plots make the three-condition comparison explicit for the free, arm-only, and torso-only settings.}
% \label{fig:condition_feature_profiles}
% \end{figure*}

To address the second research question, we examined how affective touch dynamics varied across the free, arm-only, and torso-only conditions by quantifying how strongly each feature changed across emotions within each condition. For each feature, we computed an emotion effect size using eta-squared,
\begin{equation}
\eta_f^2 = \frac{SS_{\mathrm{between}}}{SS_{\mathrm{total}}}
\end{equation}
where $SS_{\mathrm{between}}$ denotes the between-emotion sum of squares for feature $f$, and $SS_{\mathrm{total}}$ denotes the total sum of squares. Modality-level contributions within each condition were summarized using the distribution of per-feature $\eta^2$ values, and differences between modality distributions were evaluated with two-sided Mann–Whitney $U$ tests. In the free-touch condition, the analysis included body-location, pressure, and motion features; in the arm-only and torso-only conditions, it included pressure and motion features only.

\begin{table}[h]
\centering
\caption{Modality-level emotion contribution by condition, summarized from per-feature $\eta^2$ values. The free condition compares position, pressure, and motion; constrained conditions compare pressure and motion only.}
\label{tab:rq2_modality_contribution}
\begin{tabular}{lcccc}
\toprule
Condition & Position & Motion & Pressure & Key test \\
\midrule
Free & 0.049 & 0.068 & 0.057 & all $p \ge 0.248$ \\
Arm & -- & 0.038 & 0.018 & $W=36.0$, $p=0.002$ \\
Torso & -- & 0.054 & 0.084 & $W=5.0$, $p=0.041$ \\
\bottomrule
\end{tabular}
\end{table}

The effect size across body-location summaries showed that body region contributed as strongly as pressure and motion features to affective expression (Table~\ref{tab:rq2_modality_contribution}). The values in Table~\ref{tab:rq2_modality_contribution} are median per-feature $\eta^2$ within each modality, so larger values indicate stronger emotion-related variation. In the free-touch condition, body location ($\eta^2_{\mathrm{M}}=0.053$, $\eta^2_{\mathrm{Mdn}}=0.049$) and motion ($\eta^2_{\mathrm{M}}=0.067$, $\eta^2_{\mathrm{Mdn}}=0.068$) showed slightly larger effects than pressure ($\eta^2_{\mathrm{M}}=0.051$, $\eta^2_{\mathrm{Mdn}}=0.057$), but these differences were not significant (all Mann--Whitney $p \ge 0.05$), indicating no dominant modality. In the arm-only condition, motion features showed significantly larger effects than pressure features, with $\eta^2_{\mathrm{Mdn}}=0.038$ for motion and $\eta^2_{\mathrm{Mdn}}=0.018$ for pressure ($W=36.0$, $p=0.005$). In the torso-only condition, the pattern reversed, with pressure features showing larger effects than motion features, with $\eta^2_{\mathrm{Mdn}}=0.084$ for pressure and $\eta^2_{\mathrm{Mdn}}=0.054$ for motion ($W=5.0$, $p=0.045$). These results indicate that the dominant modality depends on body region: motion dominates in arm-constrained expression, whereas pressure dominates in torso-constrained expression.

To test whether these condition differences were statistically reliable, we fitted linear mixed-effects models separately for each representative feature within each condition,
\begin{equation}
\mathrm{Feature}_{ij} = \beta_0 + \sum_{k=1}^{7}\beta_k\,\mathrm{Emotion}_{ijk} + u_j + \epsilon_{ij},
\end{equation}
where \textit{Emotion} was treated as an eight-level categorical fixed effect, $u_j$ was a participant-specific random intercept, and $\epsilon_{ij}$ was the residual term. To test whether the emotion profile itself changed between constrained body regions, we then fitted interaction models to the arm-only and torso-only trials,
\begin{equation}
\begin{aligned}
\mathrm{Feature}_{ij} &= \beta_0 + \beta_1\,\mathrm{Cond.}_{ij}
+ \sum_{k=1}^{7}\beta_k\,\mathrm{Emot.}_{ijk} \\
& + \sum_{k=1}^{7}\beta^{*}_{k}\,\mathrm{Cond.}_{ij}\times\mathrm{Emot.}_{ijk}
+ u_j + \epsilon_{ij}.
\end{aligned}
\end{equation}

\begin{figure*}[h]
\centering
\includegraphics[width=0.92\textwidth]{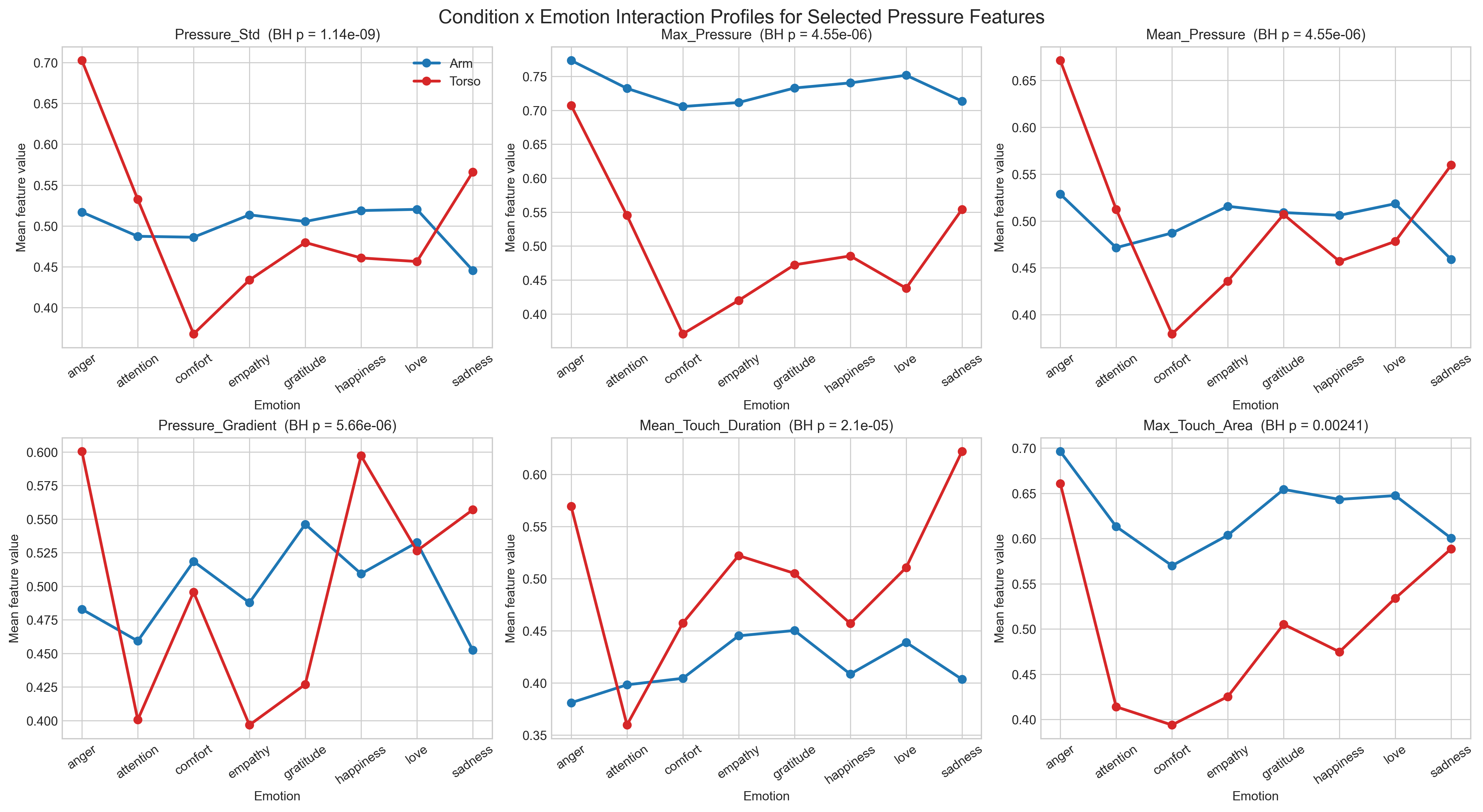}
\caption{Condition $\times$ emotion interaction profiles for the six pressure features, all of which remained significant after correction. The lines show condition-specific mean feature values across emotions for arm-only and torso-only trials.}
\label{fig:interaction_profiles}
\end{figure*}

{\sloppy
The mixed-effects models confirmed that these differences were statistically reliable. All six motion features showed significant emotion effects in all three conditions after BH correction: \textit{Mean Spectral Centroid}, \textit{Mean Peak Energy}, \textit{Mean Temporal Centroid}, \textit{Max Peak Energy}, \textit{Mean Total Energy}, and \textit{Mean Energy}. The six pressure features were more condition-dependent. In the free-touch condition, significant emotion effects were observed for \textit{Max Touch Area}, \textit{Pressure Gradient}, \textit{Pressure Std}, and \textit{Max Pressure}. In the arm-only condition, the significant pressure features were Mean Touch Duration, Pressure Gradient, and Pressure Std. In the torso-only condition, all six pressure features were significant: \textit{Max Touch Area}, \textit{Mean Touch Duration}, \textit{Pressure Gradient}, \textit{Pressure Std}, \textit{Mean Pressure}, and \textit{Max Pressure}. The duration feature highlighted this asymmetry especially clearly: \textit{Mean Touch Duration} was not significant in free touch ($p_{\mathrm{BH}}=.521$), but it was significant in both arm-only ($p_{\mathrm{BH}}=.013$) and torso-only touch ($p_{\mathrm{BH}}<.001$). The interaction models sharpened this interpretation. All six pressure features showed a significant \textit{Condition $\times$ Emotion} interaction between arm-only and torso-only touch: \textit{Pressure Std}, \textit{Max Pressure}, \textit{Mean Pressure}, \textit{Pressure Gradient}, \textit{Mean Touch Duration}, and \textit{Max Touch Area}. By contrast, none of the  motion features showed a significant interaction after correction. In other words, the emotional profile of pressure was reconfigured by body region, as shown in the Fig.~\ref{fig:interaction_profiles}, whereas the emotional profile of motion remained more nearly parallel across arm and torso touch.
\par
}

\begin{figure*}[t]
\centering
\includegraphics[width=\textwidth]{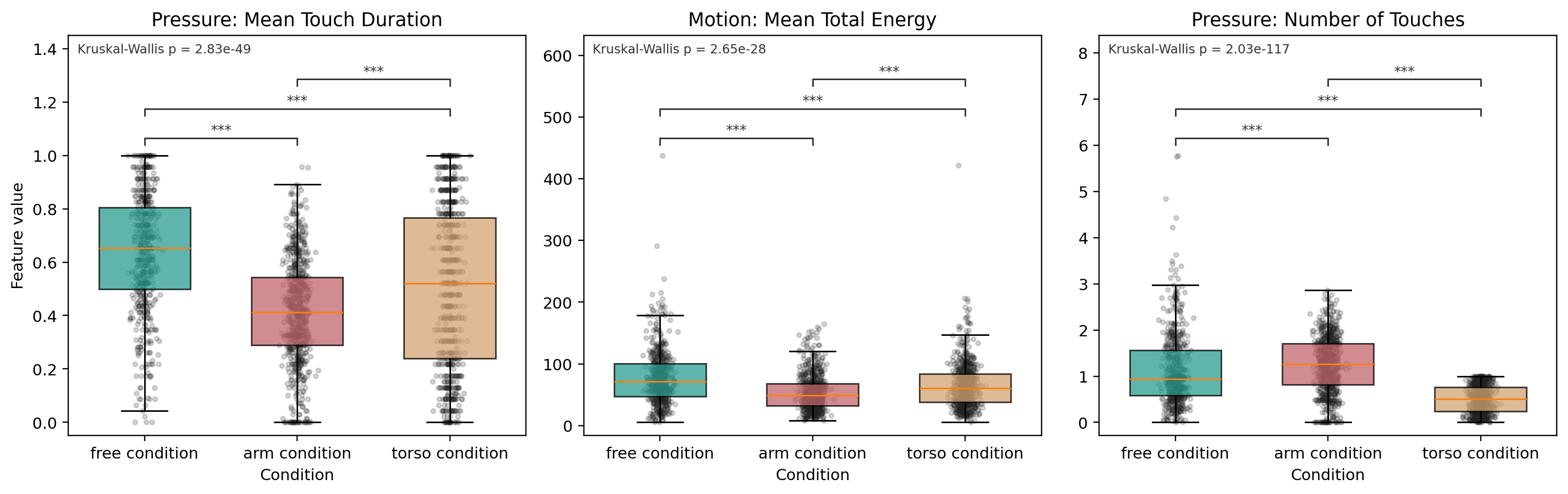}
\caption{Feature distributions across the three experimental conditions. These plots illustrate how several pressure and motion variables shift across the free, arm-only, and torso-only conditions. Pairwise significance brackets show Holm-corrected Mann--Whitney comparisons, and the inset text reports the omnibus Kruskal-Wallis $p$ value for each feature.}
\label{fig:feature_distributions}
\end{figure*}

Figure~\ref{fig:feature_distributions} provides concrete examples of how these condition differences appear in the data. \textit{Mean Touch Duration} tends to be higher in the torso-only condition than in the free or arm-only conditions, suggesting longer sustained contact when touch is restricted to the torso. Feature \textit{mean total energy} is generally highest in the free condition and lowest in the arm-only condition, consistent with stronger overall movement energy when participants can choose where to touch. \textit{Num Touches} is typically highest in the arm-only condition and lowest in the torso-only condition, suggesting that arm-only touch involves more segmented contacts whereas torso-only touch involves fewer but more sustained contacts. These descriptive distributions align with the interaction and mixed-effects results by showing how specific variables plausibly support the broader condition-dependent patterns.

% % \RQ3

\subsection{Cross-condition similarity between free-touch region use and constrained conditions}

To test whether touching a given body region in the free condition resembles the corresponding constrained condition, we computed participant- and emotion-matched centroid distances in reduced feature spaces. Specifically, we compared free-arm trials against arm-only and torso-only trials, and free-torso trials against torso-only and arm-only trials. 

To test more directly whether free-touch use of a given body region reproduced the corresponding constrained strategy, we compared free-arm trials with arm-only trials and free-torso trials with torso-only trials using paired multivariate Hotelling's $T^2$ tests on the six-feature pressure and six-feature motion vectors. The overall feature vectors differed significantly between free-arm and arm-only touch in both the pressure domain ($T^2=212.67$, $F(6,149)=34.29$, $p<0.001$) and the motion domain ($T^2=175.05$, $F(6,149)=28.23$, $p<0.001$). Free-torso and torso-only touch also differed significantly in the pressure domain ($T^2=49.84$, $F(6,50)=7.55$, $p<0.001$), whereas the motion features comparison was not significant ($T^2=12.69$, $F(6,50)=1.92$, $p=0.095$). To explore whether paired expressions using the same body region retained similar feature profiles between the free-touch and constrained conditions, we computed Pearson correlations between the paired six-feature vectors for the corresponding free-region and constrained-condition trials. The results indicate that free-arm and arm-only showed a modest but significant correlation on pressure features (mean $r=.272$, $p<0.001$), whereas free-torso and torso-only did not (mean $r=0.027$, $p=0.627$). Together, these results indicate that free-touch behaviour did not simply collapse onto the corresponding constrained condition. Instead, constrained interactions preserved only partial structure while reorganizing the overall embodied strategy.

\subsection{Questionnaire measurements}

\subsubsection{Scale reliability and pre–post effects}

The preceding analyses showed that affective touch strategies were strongly shaped by embodiment: participants adapted both where they touched and how they touched across free and constrained conditions, and these strategies did not transfer directly across body regions. An additional question is whether affective touch interactions influence how people perceive the robot, even when the interaction is one-directional and task-oriented.

To examine how tactile interaction influenced participants’ perceptions of the robot, we analyzed \textit{Agency \& Experience} (8 items), \textit{Agency \& Communality} (11 items), and the \textit{IOS} scale. Internal consistency for \textit{Agency \& Experience} and \textit{Agency \& Communality}, was assessed using Cronbach’s $\alpha$, following standard practice for reliability evaluation \cite{cronbach1951coefficient}. Both scales demonstrated good internal consistency in the pre-interaction and post-interaction phases (\textit{Agency \& Experience}: $\alpha_{pre}=0.905$, $\alpha_{post}=0.908$; \textit{Agency \& Communality}: $\alpha_{pre}=0.899$, $\alpha_{post}=0.934$). Pre–interaction and post-post differences were tested using Wilcoxon signed-rank tests. No significant change was observed for \textit{Agency \& Experience}, ($M_{pre}=3.05$, $M_{post}=3.05$, $p=0.629$), or for \textit{Agency \& Communality}, ($M_{pre}=4.24$, $M_{post}=4.27$, $p=0.745$). In contrast, the \textit{IOS} measure showed a significant increase, ($M_{pre}=3.13$, $M_{post}=3.87$, $p=.002$, $d_z=0.64$). These results suggest that affective tactile interaction did not significantly alter broader perceptions of agency or communality, but did increase perceived relational closeness toward the robot.

\begin{figure*}[t]
\centering
\includegraphics[width=0.96\textwidth]{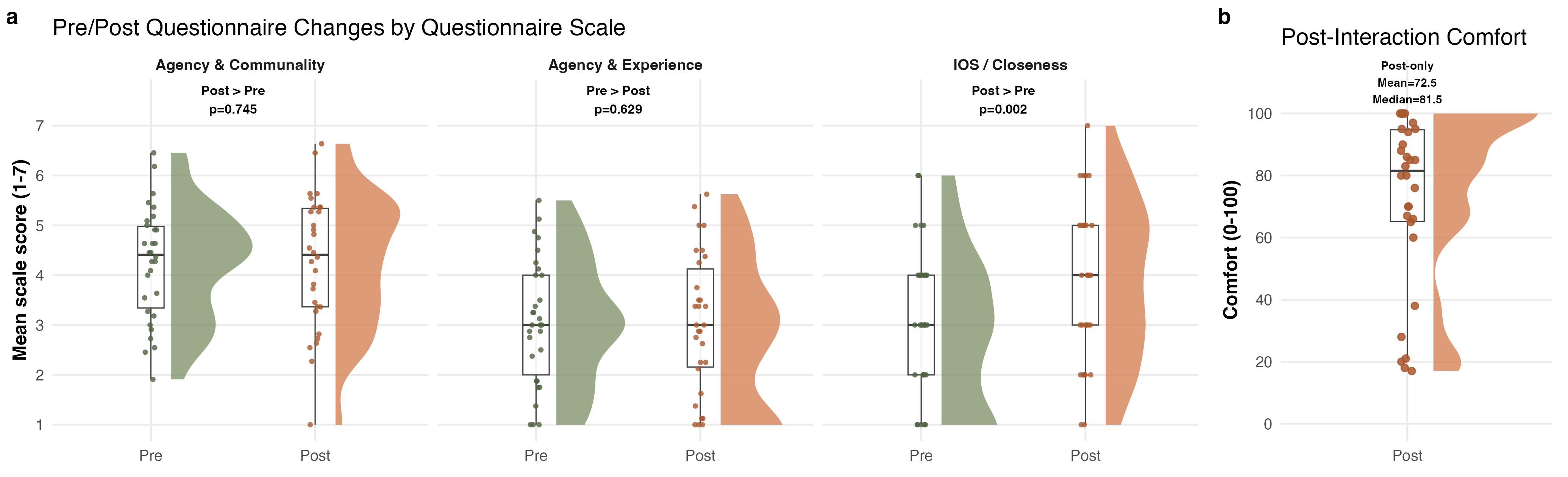}
\caption{Raincloud-style plots for before and after interaction self report rating changes. Each panel combines a half violin density, a narrow boxplot, and participant-level points. Labels report the direction of change and the Wilcoxon $p$ value. The two multi-item scales showed good internal consistency (\textit{Agency \& Experience}: $\alpha_{pre}=0.905$, $\alpha_{post}=0.908$; \textit{Agency \& Communality}: $\alpha_{pre}=0.899$, $\alpha_{post}=0.934$). The clearest pre/post effect was in the single-item IOS scale.}
\label{fig:questionnaire_family}
\end{figure*}

\begin{figure*}[t]
\centering
\includegraphics[width=0.84\textwidth]{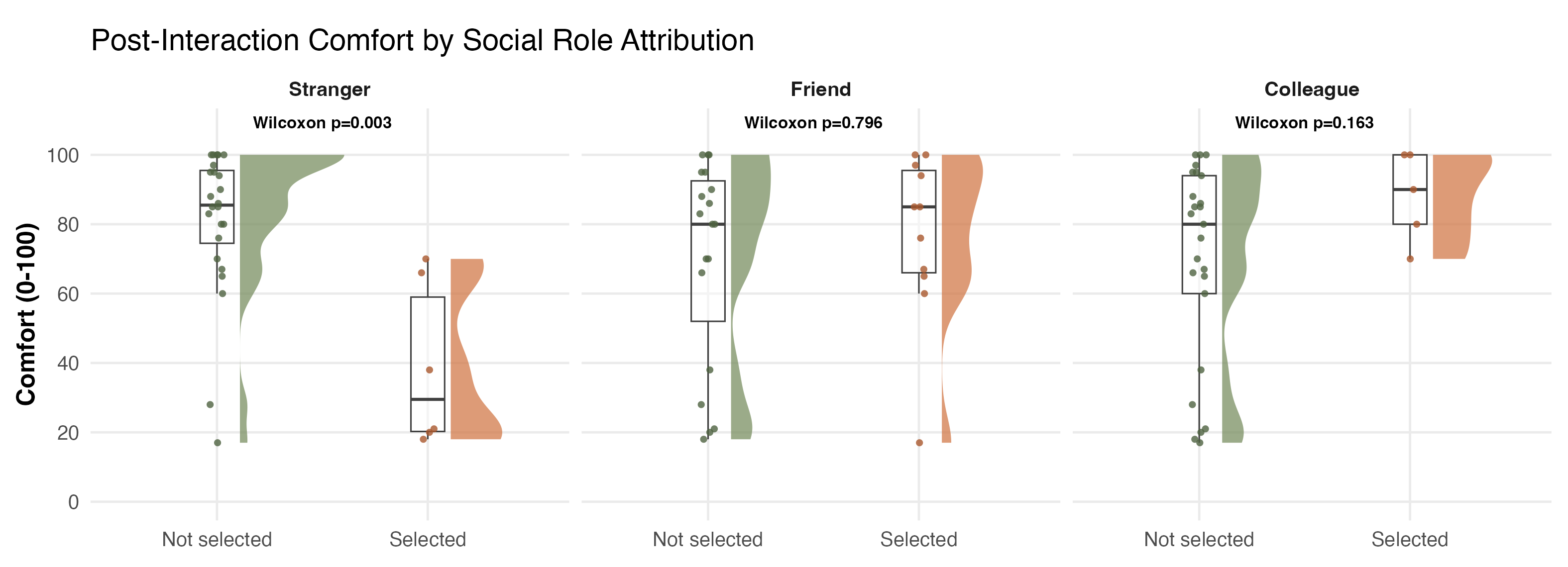}
\caption{Post-interaction comfort scores split by social-role attribution. Each panel shows a half violin, a narrow boxplot, and participant-level points for participants who did and did not endorse the role in the post questionnaire. The clearest separation was for \textit{Stranger}: participants who still described iCub as a stranger reported much lower comfort than those who did not.}
\label{fig:questionnaire_comfort_roles}
\end{figure*}

\subsubsection{Comfort as a function of social-role attribution}

Finally, post-interaction comfort ratings indicated that the touch interaction was generally experienced as comfortable (M = 72.47/100, Mdn = 81.50/100). We examined how comfort related to participants’ social framing of the robot, as shown in Fig.~\ref{fig:questionnaire_comfort_roles}, using correlations and Wilcoxon rank-sum tests for selected versus non-selected role groups. Comfort showed its strongest association with post-interaction IOS scores (Pearson $r=0.41$, $p=0.026$), with a marginal trend for \textit{Agency \& Communality} ($r=0.36$, $p=0.053$), while associations with \textit{Agency \& Experience} were weaker. The most pronounced effect emerged at the level of social-role attribution. Participants who characterised iCub as a \textit{stranger} reported substantially lower comfort (M = 38.83) compared to those who did not (M = 80.88; Wilcoxon $p=0.003$), as illustrated in Fig.~\ref{fig:questionnaire_comfort_roles}. In contrast, \textit{friend} and \textit{colleague} labels were not associated with significant differences in comfort. Taken together, these findings indicate that comfort is most strongly aligned with participants’ social framing of the robot, particularly at the negative end: participants who maintained a \textit{stranger} interpretation experienced markedly lower comfort during touch interaction. 

% In addition, higher self-reported familiarity with robots was associated with greater perceived closeness (Spearman $\rho=0.43$, $p=0.019$).

\subsubsection{Shifts in perceived social roles}

Perceived social role shifts revealed a small but interpretable qualitative pattern. Among the 30 matched participants, 9 changed their perceived social-role attribution from pre- to post-interaction, whereas 21 retained the same role profile. Two participants initially identified the robot as a colleague or student and subsequently reclassified it as a relative following tactile interaction. Six participants shifted from more distant roles (neighbour, relative, or stranger) to friend, and one participant changed from stranger to neighbour. Although most participants did not alter their role attribution, the observed transitions predominantly moved toward more affiliative interpretations of the robot, with a notable convergence toward the friend category.

\section{Discussion}

% \subsection{RQ1: What principles of tactile sensor placement and coverage are required for robots to support affective communication?}
% Write the comparison with humans here as well, maybe.
% \cite{suvilehto2015topography}

Our results suggest three design principles. First, tactile coverage should prioritize socially salient upper-body regions rather than merely maximizing nominal skin coverage. The free-touch data showed concentrated use of the upper arms, torso, hands, exposed joints, and face/head, whereas the legs, feet, pelvic region, and gluteal region were almost never selected. This pattern is majorly consistent with psychological maps of socially acceptable touch, which show that affective touch is unevenly distributed across the body and concentrated in regions with stronger social meaning \cite{suvilehto2023and}\cite{suvilehto2015topography}; however, prior work by Suvilehto \textit{et al.} \cite{suvilehto2015topography} shows that in human–human interaction, arms and hands are generally acceptable touch regions across different relationships, such as friends and strangers, while the torso is usually less permissible. In contrast, our results show that participants more often touched the robot’s torso. This suggests that social norms from human–human touch do not directly transfer to human–robot interaction, and people may feel more comfortable interacting with robot body regions that are typically restricted in human contexts. In addition, robot's body and appearance may influence touch behaviour. Prior work has shown that perceived social roles, including inferred gender and relational expectations, shape where and how people feel comfortable touching others \cite{suvilehto2015topography}. This effect may extend to human–robot interaction, where different robot forms evoke distinct interaction scripts. For example, in a study with the Nao robot, which has a child-like form and is approximately half the size of iCub, participants frequently touched the arms but rarely the torso \cite{andreasson2018affective}. Such patterns suggest that touch behaviour is not determined solely by emotion but also by how users perceive the robot’s body and social role. In Andreasson \textit{et al.}'s setup, the robot was positioned on a table rather than fixed to a base, which may have further constrained touch behaviour due to perceived instability or risk of tipping. Together, these factors indicate that embodiment-related constraints and perceptions likely contribute to the observed distribution of touch across body regions. Second, sensor coverage should include transition zones rather than only canonical body parts. The frequent use of exposed joints indicates that people do not divide the robot body into neat engineering segments; they use mechanically visible, reachable, and socially interpretable regions as expressive contact zones. Taken together, these findings answer the first research question: tactile sensor placement for affective communication should privilege upper-body regions with social salience and natural reachability, should preserve information about where touch occurs, and should not treat sensor placement as a purely technical optimization problem divorced from human social touch behaviour.

% \subsection{RQ2: How do touch dynamics vary by body region, and what does this reveal about embodied constraints in social robots?}

The second research question is directly answered by the different modality comparison analyses and the interaction models. Touch dynamics varied by body region in two distinct ways. First, under free-touch conditions, participants relied strongly on 
where they touched, with pressure and motion providing useful information as well. This suggests that when spatial freedom is available, affective intent is partly expressed by body-region choice itself. Second, once location was constrained, the dominant dynamic channel shifted with body region. In the arm-only condition, motion features were more informative for emotion recognition than pressure features, indicating greater variation across emotional expressions. However, participants applied higher overall pressure in the arm-only condition than in the torso-only condition. This may be because social constraints associated with the torso, as well as concerns about potentially damaging the robot, discouraged participants from applying high overall pressure. However, participants still varied the applied pressure across emotional expressions. This pattern may also be influenced by sensor limitations, as pressures beyond a certain level may approach the sensor’s maximum range, reducing sensitivity to further increases. In the torso-only condition, the opposite pattern held, pressure features were more informative than motion features while higher motion energy baseline was obtained. The mixed-effects models showed that emotional variation in motion was reliable in all three conditions, but pressure was selectively stronger in torso and free touch than in arm touch.

Our results also indicate that the significant interactions were concentrated in pressure features, not motion features. This means that the emotional pressure profile is not simply stronger or weaker across body regions; it is reorganized by body region. In contrast, motion appears to function more as a broadly available affective channel whose emotional structure is relatively consistent across arm and torso. This distinction is theoretically meaningful. It aligns with embodied and dynamic views of emotion, in which bodily affordances shape not only whether an action is possible, but also which expressive channel becomes most effective in a given spatial context. The arm affords lateral movement, rhythmic brushing, and repeated motion, whereas the torso affords larger contact surfaces, sustained contact, and pressure-based modulation. Our data therefore, support the claim that affective touch is body-region dependent in a strong sense: different robot body regions invite different expressive strategies, and those strategies cannot be assumed to transfer unchanged across the body.

% \subsection{RQ3: How does task-oriented, one-directional affective touch toward a robot influence people’s attitudes and impressions of the robot?}

The questionnaire analyses now allow a direct answer to the third research question. Our results shown that there is a clear attitudinal effect of the interaction, which was an increase in relational closeness rather than a broad rise in anthropomorphic attributions. However, the only clear significant change was found in the IOS measure, whereas the \textit{Agency \& Experience} and \textit{Agency \& Communality} remained stable despite good internal consistency. This indicates that even task-oriented, one-directional tactile interaction can increase perceived closeness between the robot and participants, while agency is not readily influenced by such interaction. Agency likely requires more than brief, unidirectional contact, and may depend on sustained, reciprocal interaction, perceived responsiveness, and opportunities for participants to attribute intentionality and control to the robot over time \cite{wen2022sense}. This pattern suggests that one-directional affective touch does influence how participants position the robot socially, but it does so more strongly at the level of felt relational distance than at the level of explicit belief about the robot's inner capacities. Participants came to feel somewhat closer to iCub and were somewhat more willing to attribute friend-like roles to it, yet they did not uniformly become more convinced that the robot could feel, understand, or act as a fully sentient social partner. The participant-level role changes are consistent with this interpretation: only 9 of the 30 matched participants changed the role they attributed to the robot at all, but when change occurred it most often moved in a more affiliative direction, especially toward \textit{friend}.

In addition, participants generally reported that touching iCub was comfortable, even though the task remained one-directional and the robot provided no contingent tactile reciprocity. At the same time, comfort was not socially neutral, it was higher in participants who rated iCub as friendlier and closer, and much lower in those who still categorized iCub as a \textit{stranger}. This means that the interaction was not experienced as aversive or inappropriate overall, but comfort alone was not sufficient to produce large across-the-board changes in anthropomorphic belief. In conclusion, one-directional affective touch can make the robot feel relationally closer and socially more approachable, but it does not by itself fully establish the robot as an emotionally capable reciprocal partner.

Prior work implicitly assumes invariant affective features, relies on single-region datasets, and largely ignores embodiment \cite{albawi2018social}\cite{jung2015touch}\cite{ren2024touch}. These limitations arise from treating touch as a context-free signal rather than an interactional process grounded in the body and situation. The present findings instead point to three implications. Models need to account for body region. Training data needs to cover a wider range of spatial locations. Multimodal fusion needs to adapt to the relevance of each affective dimension. Taken together, the results suggest moving away from fixed mappings between signals and emotions. Affective meaning is shaped through the interaction between tactile dynamics, body location, and situational context. This view supports the development of affective HRI systems that are more robust and generalisable across users, contexts, and interaction settings.

\subsection{Limitations}

Several limitations should be considered when interpreting these findings. First, although the iCub provides relatively rich full-body tactile coverage, its physical morphology, posture, and accessibility constraints may have shaped participants’ behaviour. In this study, the iCub robot has a child-like body shape, which may influence how participants interact with it. This could partially explain the higher frequency of touches to the torso, a region that is less commonly targeted in human–human tactile interaction studies. In addition, the robot’s existing ''skin'' may have influenced touch behaviour, potentially shaping participants’ interaction patterns, even at an implicit level. For example, one participant interpreted the skin as clothing and asked whether touch should be restricted to these areas. However, our results show that participants also interacted with regions such as joints and the face, indicating that the presence of the skin did not fully constrain expressive behaviour when no restrictions were imposed. This suggests that, although physical appearance can guide interaction, participants are willing to express freely beyond skin boundaries when explicitly instructed that they don't have any limitations for expression. In real-world settings, where such instructions are absent, these factors may limit expressive touch, people might adapt to the robot's sensing ability to make sure that robot can perceive their touch expression.

The robot was in a fixed upright position, and participants were seated and interacted with the robot while it remained in a standing posture, which likely biases touch toward easily reachable upper-body regions. The near absence of contact with lower-body areas (e.g., legs, pelvic, and gluteal regions) should therefore not be interpreted purely as a lack of affective relevance, but also as a consequence of physical reachability, social norms, and interaction ergonomics. Different robot postures, such as seated or lying configurations, or different participant positions, may lead to different spatial touch distributions. Beyond the static configuration used in the main study, robot movement is also likely to affect interaction behaviour. In particular, participants may avoid contacting exposed joints (e.g., shoulders) when the robot is in motion due to perceived safety concerns. This contrasts with human–human interaction, where shoulder tapping is commonly used to attract attention and is generally not associated with safety concerns. Second, the interaction was one-directional and task-oriented. The robot did not provide contingent tactile or behavioural feedback, which limits conclusions about reciprocal affective touch and long-term relationship formation. More interactive paradigms may produce stronger or qualitatively different attitudinal effects. Third, the participant sample was relatively homogeneous and culturally localized, which may limit generalizability. Although prior work suggests cross-cultural consistency in touch norms, subtle cultural or demographic differences may still influence touch strategies. In addition, this study is based on a one-shot interaction, and longer-term relationships with the robot may significantly shape how and where people touch the platform. Participants’ perceived social roles of the robot may also differ along the time. For example, one participant expressed a desire to have the iCub at home and interact with it over time, while some participants were already getting bored with this one-directional affective touch expressions. Such variation may suggest that some participants may treat the robot as a functional device, while others may perceive it as a child-like entity requiring care. As interaction duration increases, these perceptions may evolve, potentially leading to changes in touch behaviour. Future work should therefore examine how perceived social roles and emotional bonding develop over time and influence interaction patterns. Finally, the selected feature set prioritised interpretability over completeness. While representative pressure and motion features were analysed, more complex spatiotemporal patterns may further refine understanding of affective touch dynamics.

\section{Conclusion}

This study investigated how emotional content is expressed across different body locations of the robot, as well as how spatial constraints influence these expressions. It extends prior work that has largely focused on gestures and interactions limited to specific body regions; our results demonstrate that affective touch is fundamentally body region dependent, both in spatial selection and in dynamic execution. Participants preferentially engaged socially accessible upper-body regions, while less frequently used regions exhibited stronger emotion-specific selectivity, revealing a dissociation between interaction frequency and affective informativeness.

The analysis further shows that expressive mechanisms reorganize under spatial constraints. In free interaction, affective intent is partly conveyed through body region choice, whereas constrained interaction shifts expression toward region specific dynamics, with motion dominating in arm based interaction and pressure dominating in torso based interaction. Importantly, touch strategies did not transfer directly between free and constrained conditions, indicating that affective touch cannot be modelled as a location invariant signal. At the perceptual level, affective touch increased perceived relational closeness without significantly altering attributed agency, suggesting that tactile interaction primarily modulates relational experience rather than explicit cognitive beliefs.

These findings have direct implications for affective computing systems. Models for touch based emotion recognition and generation should explicitly condition on body region, incorporate spatially diverse training data, and adopt adaptive representations of pressure and motion. More broadly, affective touch should be treated as an embodied, context sensitive communication process rather than a fixed set of transferable gestures.

\balance

\bibliographystyle{ieeetr}
\bibliography{references}

\vfill

\end{document}